\documentclass{article}

\usepackage{microtype}
\usepackage{graphicx}
\usepackage{subcaption}
\usepackage{booktabs} 

\usepackage{hyperref}

\usepackage[preprint]{icml2026}
\makeatletter
\renewcommand{\ICML@preprint}{}
\makeatother

\usepackage{amsmath}
\usepackage{amssymb}
\usepackage{mathtools}
\usepackage{amsthm}

\usepackage[table]{xcolor}
\usepackage{xspace}
\usepackage[most]{tcolorbox}
\usepackage{eurosym}
\usepackage{enumitem}
\usepackage{multirow}
\definecolor{customgreen}{RGB}{0,128,0}
\definecolor{seablue}{RGB}{0, 84, 147}
\definecolor{bgblue}{RGB}{225, 240, 255}

\newcommand{\ours}{RuCL\xspace}

\usepackage[capitalize,noabbrev]{cleveref}

\theoremstyle{plain}

\theoremstyle{definition}

\theoremstyle{remark}

\usepackage[textsize=tiny]{todonotes}

\icmltitlerunning{RuCL: Stratified Rubric-Based Curriculum Learning for Multimodal Large Language Model Reasoning}

\begin{document}

\twocolumn[
  \icmltitle{RuCL: Stratified Rubric-Based Curriculum Learning for \\ Multimodal Large Language Model Reasoning}



  \icmlsetsymbol{equal}{*}

  \begin{icmlauthorlist}
    \icmlauthor{Yukun Chen}{equal,siat,ucas}
    \icmlauthor{Jiaming Li}{equal,siat,ucas}
    \icmlauthor{Longze Chen}{siat,ucas}
    \icmlauthor{Ze Gong}{siat}
    \icmlauthor{Jingpeng Li}{alibaba}
    \icmlauthor{Zhen Qin}{alibaba}
    \icmlauthor{Hengyu Chang}{alibaba}
    \icmlauthor{Ancheng Xu}{siat,ucas}
    \icmlauthor{Zhihao Yang}{siat,ucas}
    \icmlauthor{Hamid Alinejad-Rokny}{hamid}
    \icmlauthor{Qiang Qu}{siat}
    \icmlauthor{Bo Zheng}{alibaba}
    \icmlauthor{Min Yang}{siat}
  \end{icmlauthorlist}

  \icmlaffiliation{siat}{Shenzhen Institute of Advanced Technology, Chinese Academy of Sciences}
  \icmlaffiliation{ucas}{University of Chinese Academy of Sciences}
  \icmlaffiliation{alibaba}{Alibaba Group}
  \icmlaffiliation{hamid}{School of Biomedical Engineering, UNSW Sydney}
  
  \icmlcorrespondingauthor{Ze Gong}{ze.gong@siat.ac.cn}
  \icmlcorrespondingauthor{Bo Zheng}{bozheng@alibaba-inc.com}
  \icmlcorrespondingauthor{Min Yang}{min.yang@siat.ac.cn}

  \icmlkeywords{Machine Learning, ICML}

  \vskip 0.3in
]



\printAffiliationsAndNotice{\icmlEqualContribution}  

\begin{abstract}
  Reinforcement Learning with Verifiable Rewards (RLVR) has emerged as a prevailing paradigm for enhancing reasoning in Multimodal Large Language Models (MLLMs).
  However, relying solely on outcome supervision risks reward hacking, where models learn spurious reasoning patterns to satisfy final answer checks.
  While recent rubric-based approaches offer fine-grained supervision signals, they suffer from high computational costs of instance-level generation and inefficient training dynamics caused by treating all rubrics as equally learnable.
  In this paper, we propose \textbf{Stratified Rubric-based Curriculum Learning (\ours)}, a novel framework that reformulates curriculum learning by shifting the focus from data selection to reward design. 
  \ours generates generalized rubrics for broad applicability and stratifies them based on the model's competence. By dynamically adjusting rubric weights during training, \ours guides the model from mastering foundational perception to tackling advanced logical reasoning. 
  Extensive experiments on various visual reasoning benchmarks show that \ours yields a remarkable $\textbf{+7.83\%}$ average improvement over the Qwen2.5-VL-7B model, achieving a state-of-the-art accuracy of $\textbf{60.06\%}$.
\end{abstract}

\section{Introduction}

Multimodal Large Language Models (MLLMs) have demonstrated remarkable capabilities in complex visual reasoning tasks, spanning from mathematical problem-solving to chart understanding~\citep{yao2024mulberry,liu2025othink,peng2025skywork,amizadeh2020neuro,garcez2019neural}. 
To further augment these reasoning capabilities, Reinforcement Learning with Verifiable Rewards (RLVR)~\citep{shao2024deepseekmath, cui2025process, li2025implicit} has emerged as a prevalent post-training paradigm.
By employing straightforward rule-based verification, RLVR avoids the reliance on costly reward models~\citep{meng2025mm,liu2025noisyrollout,xu2025mixed}.

However, this outcome-based reward mechanism suffers from a fundamental limitation: it overemphasizes final answer correctness at the expense of intermediate reasoning quality.
As a result, models are prone to learning spurious reasoning patterns or exploiting superficial shortcuts.
This frequently leads to the generation of contradictory or hallucinatory intermediate steps that serendipitously arrive at correct answers.
Such ``reward hacking'' phenomenon severely compromises the reliability of the reasoning.

\begin{figure}[!t]
	\centering
	\includegraphics[width=0.9\columnwidth]{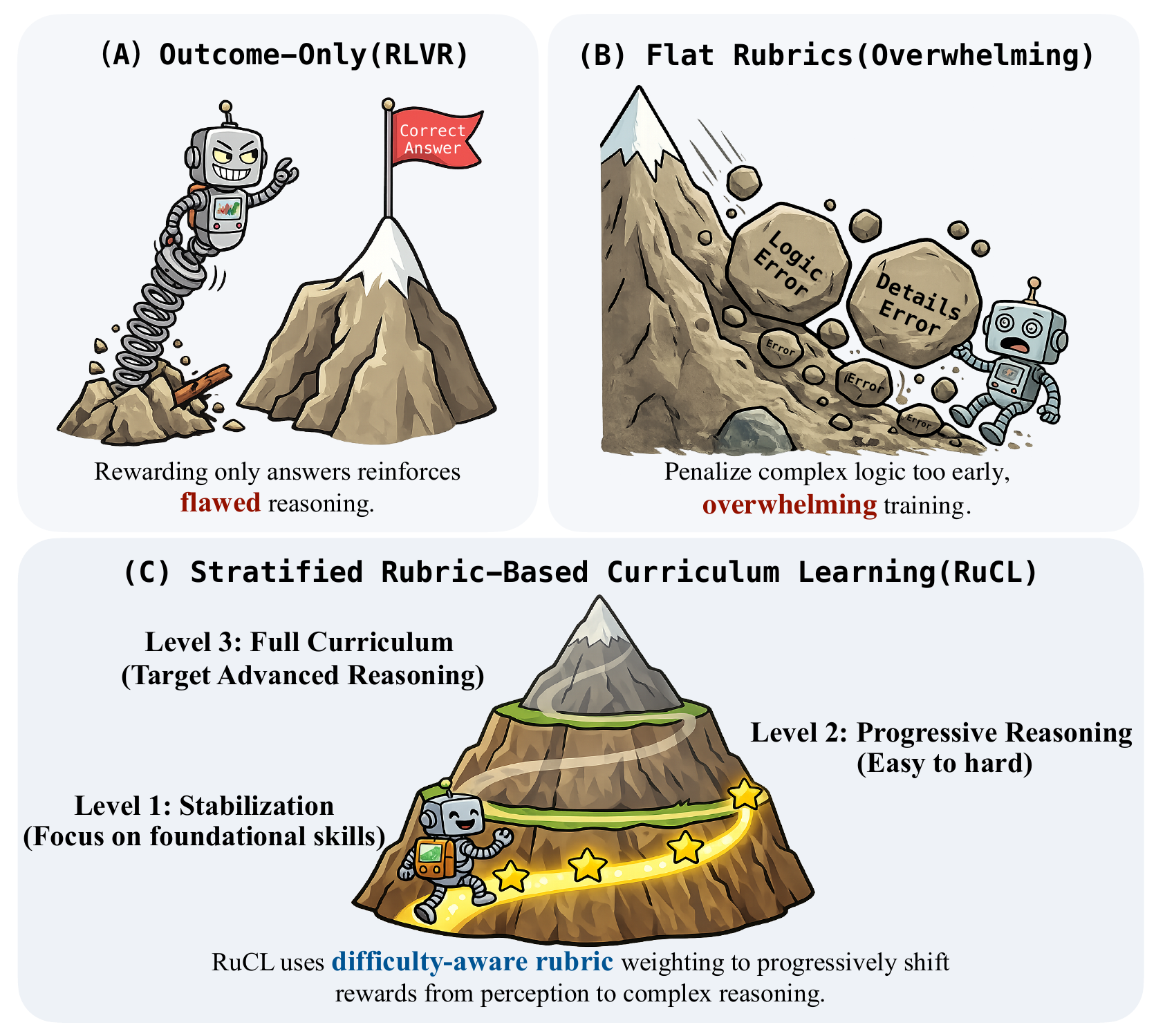}
    \caption{\textbf{Comparison of reward paradigms.} We move beyond (A) outcome-only signals and (B) unstructured dense feedback. (C) Our \textbf{\ours} framework organizes rubrics into a stratified curriculum, aligning reward complexity with the model's progressive learning stages.}
	\label{fig:figure1}
\end{figure}

While recent LLM-as-a-Judge frameworks successfully mitigate reward hacking by constructing rubrics to assess the validity of reasoning trajectories~\citep{viswanathan2025checklists,gunjal2025rubrics}, they are hampered by two fundamental limitations~\citep{huang2025reinforcement,zhou2025breaking,pathak2025rubric}. 
First, generating rubrics at the \textit{instance level} incurs high computational overhead, especially during online reinforcement learning setting.
Second, and more importantly, existing methods treat all rubrics equally challenging throughout the training process, lacking a principled mechanism to account for heterogeneous learnability across evaluation rubrics. Consequently, models are penalized for complex logical failures before mastering basic skills such as visual perception, resulting in noisy gradient signals and hindering efficient convergence.

Drawing inspiration from Curriculum Learning (CL)~\citep{bengio2009curriculum,parashar2025curriculum}, which traditionally organizes training data from easy to hard, we propose \textbf{Stratified Rubric-based Curriculum Learning (\ours)}, a novel framework that applies curriculum learning directly to reward design rather than data selection. Instead of treating all rubrics uniformly throughout training, our key insight is to organize and schedule rubrics according to their learnability, enabling the model to acquire reasoning skills in a structured and progressive manner (Fig.~\ref{fig:figure1}).

\ours can be explained as a two-phase process: \textbf{(1) Generalized Rubric Construction and Stratification}: We adopt a data-driven approach to generate generalized rubrics that capture essential reasoning primitives shared across tasks, rather than relying on costly instance-specific evaluation. We estimate the model’s initial competence on each rubric and stratify them by empirical proficiency level, ranging from foundational skills to advanced reasoning abilities. \textbf{(2) Dynamic Curriculum Learning}: During training, \ours dynamically adjusts the weights of these rubrics based on the model's evolving capabilities. Training initially prioritizes foundational rubrics (e.g., visual element recognition). As the model demonstrates competence, the framework automatically shifts focus towards hard rubrics (e.g., complex logical deduction), effectively guiding the model from basic perception to advanced reasoning. Finally, the combination of final answer reward and rubric-based reward jointly promotes the model's reasoning capabilities.

Our contributions are summarized as follows:
\begin{enumerate}[label=(\roman*)]
    \item We introduce \ours, a reward-centric curriculum framework that dynamically aligns rubric difficulty with model competence.
    \item We instantiate \ours with a data-driven rubric construction pipeline, an applicability-aware evaluation mechanism, and a performance-triggered curriculum scheduler, yielding a practical and scalable reward design for rubric-based approaches.
    \item We conduct extensive experiments across seven benchmarks, showing that \ours achieves an average performance gain of $\textbf{7.83\%}$, and provide detailed ablation studies validating its effectiveness.
\end{enumerate}

\section{Related Work}

\textbf{Post-training for MLLMs.} 
Early MLLM reasoning methods, such as LLaVA-Reasoner~\cite{zhang2025improve}, MPO~\cite{wang2024enhancing}, and Insight-V~\cite{rafailov2023direct}, rely on rationale distillation, human preferences, or iterative DPO, but are limited by heavy supervision and low scalability. To address this, Reinforcement Learning with Verifiable Rewards (RLVR)~\cite{ma2025s,chu2025gpg} verifies final answers against ground truth, enabling scalable reasoning improvement. For example, Vision-R1~\cite{huang2025vision} leverages teacher MLLMs to generate chain-of-thought (CoT) data, DeepScaler~\cite{deepscaler2025} and Light-R1~\cite{wen2025light} combine supervised and RL training, and VL-Rethinker~\cite{wang2025vl}, SRPO~\cite{wan2025srpo}, and GThinker~\cite{zhan2025gthinker} use reflection-aware rewards. Despite these advances, sparse outcome-based rewards leave models prone to reward hacking via spurious reasoning.

\textbf{Rubrics as Rewards.}
To address the opacity and sparsity of outcome-based supervision, recent work uses structured rubrics to evaluate intermediate reasoning processes, decomposing tasks into explicit, verifiable criteria.
Rubrics have proven effective in domains such as medical reasoning~\cite{arora2025healthbenchevaluatinglargelanguage}, code generation~\cite{mahdaoui2025automated}, and instruction following~\cite{pathak2025rubric,galvan2025rubrik,fan2024sedareval,winata2025datasheets}.
LLM-as-a-Judge frameworks~\cite{team2025kimi,viswanathan2025checklists} integrate rubrics into reinforcement learning, providing more informative reward signals than standard RLVR~\cite{huang2025reinforcement,gunjal2025rubrics}. However, existing approaches typically generate instance-specific rubrics and treat all rubrics as equally learnable, lacking a principled mechanism to account for heterogeneous difficulty across reasoning skills.

\textbf{Curriculum Learning.}
Curriculum Learning (CL), introduced by~\cite{bengio2009curriculum}, organizes training into phases to mimic human learning and enable progressive skill acquisition~\cite{parashar2025curriculum,shi2025efficient,chen2025self,song2025fastcurl}. Kwai Keye-VL~\cite{team2025kwai} improves capability and stability by adopting a multi-stage training recipe that structures both pre-training and post-training, while VL-Contigo~\cite{yuan2025vl} implements an ``easy-to-hard'' RL curriculum with online difficulty weighting across three stages. These prior approaches focus on data-level curricula; in contrast, we apply CL at the rubrics level, dynamically adjusting rubric weights during RL to balance training stability and reasoning performance.

\begin{figure*}[!t]
  \centering
  \includegraphics[width=0.9\textwidth]{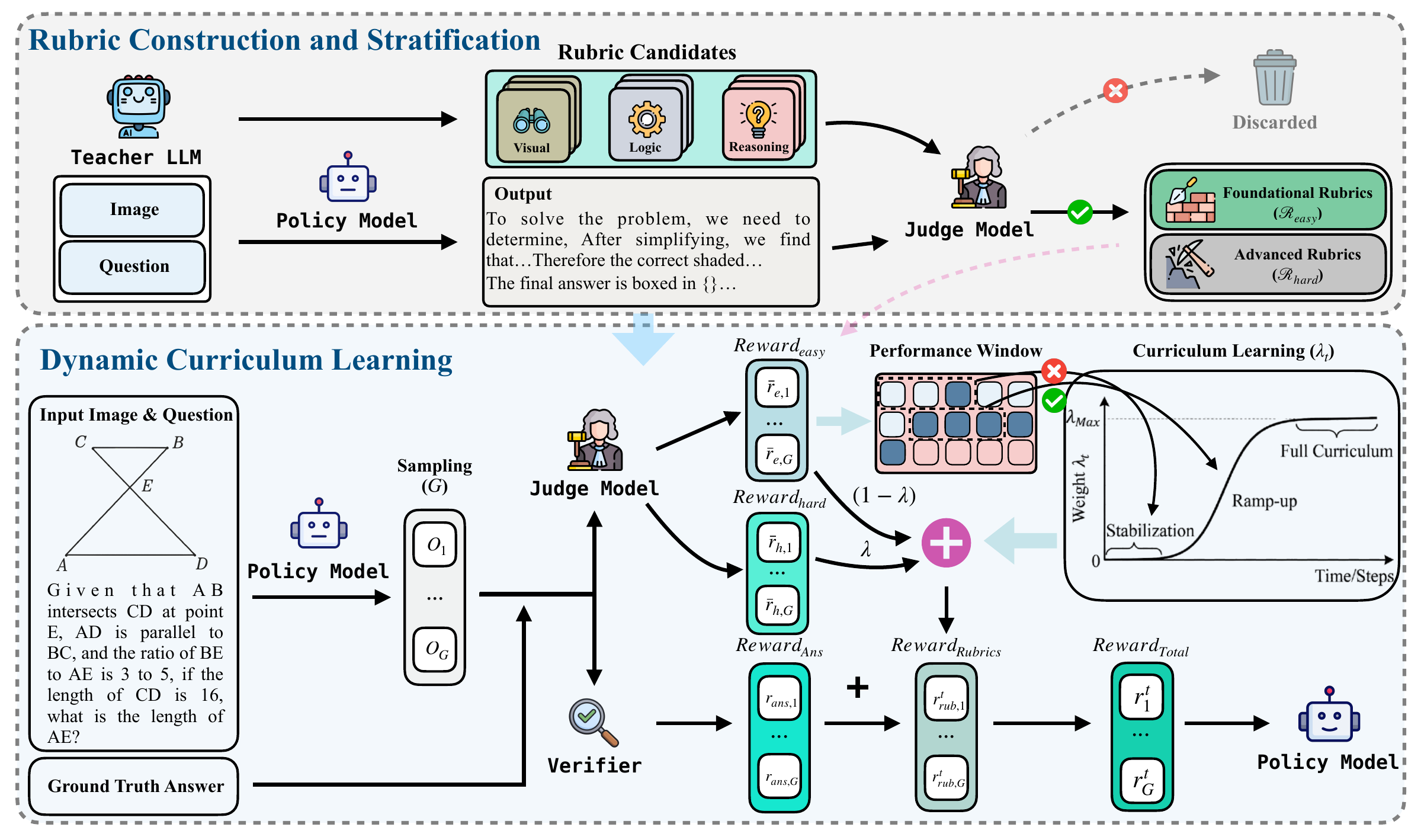}
  \caption{\textbf{Overview of Stratified Rubric-based Curriculum Learning (\ours).} 
The framework proceeds in two stages: 
(Top) \textbf{Generalized Rubric Construction and Stratification}, where evaluation rubrics are generated and categorized into Foundational ($\mathcal{R}_{\text{easy}}$) and Advanced ($\mathcal{R}_{\text{hard}}$) tiers based on empirical difficulty. 
(Bottom) \textbf{Dynamic Curriculum Learning}, where the rubric-based reward is synthesized via a dynamic weighting mechanism controlled by a scheduler. By adjusting the weight $\lambda$ based on real-time performance, \ours progressively shifts the optimization focus from mastering basic skills to tackling complex reasoning.}
  \label{fig:method}
\end{figure*}

\section{Stratified Rubric-based Curriculum Learning (RuCL)}


In this work, we focus on rubric-based rewards to improve

the reasoning capabilities of Multimodal Large Language Models (MLLMs). 
While rubrics provide fine-grained supervision over reasoning processes, existing methods typically combine them with fixed weights, ignoring differences in difficulty and learnability. This results in noisy gradients and inefficient optimization. We propose \textbf{Stratified Rubric-based Curriculum Learning (\ours)}, which applies curriculum learning directly to reward design by progressively emphasizing rubrics of increasing difficulty. In this section, we first formalize the learning objective, then detail rubric construction and the curriculum mechanism.

\begin{table*}[t]
    \caption{\textbf{The stratified reward system.} The evaluation rubrics
    are categorized by difficulty and implemented via either a generative LLM Judge or a deterministic Answer Verifier. Detailed rubric definitions and scoring criteria are provided in Appendix~\ref{app:Rubric Construction and Filtering}.}
    \centering
    \begin{small}
    \renewcommand{\arraystretch}{1.2}
        \begin{tabular}{l l c c}
            \toprule
            \textbf{Rubric Criterion} & \textbf{Evaluation Focus} & \textbf{Difficulty Stratum} & \textbf{Evaluator} \\
            \midrule
            
            \textsc{Visual Presence} & Penalizes object and attribute hallucinations. & \multirow{4}{*}{Foundational ($\mathcal{R}_{\text{easy}}$)} & \multirow{6}{*}{LLM Judge} \\
            \textsc{Entity Extraction} & Isolates the specific Region of Interest (ROI). & & \\
            \textsc{Intent Alignment} & Checks compliance with scope and constraints. & & \\
            \textsc{Conclusion Match} & Ensures the answer logically follows the reasoning. & & \\
            
            \cmidrule{1-3}
            
            \textsc{Step Coherence} & Detects logical gaps and internal contradictions. & \multirow{2}{*}{Advanced ($\mathcal{R}_{\text{hard}}$)} & \\
            \textsc{Evidence Grounding} & Validates inferences using specific visual cues. & & \\
            
            \midrule
            
            \textsc{Answer Correctness} & Verifies the final answer against the ground truth. & \centering{---} & Answer Verifier \\
            \bottomrule
        \end{tabular}
    \end{small}
    \label{tab:rubric_reordered}
\end{table*}

\subsection{Problem Formulation}

We consider a Reinforcement Learning (RL) setting.
Given an input query $x$ (e.g., an image-text pair), a policy $\pi_\theta(y \mid x)$ generates a response $y$.  
The learning signal is provided by a scalar reward function $r^{(t)}(y \mid x)$, which may vary over the training step $t$ to reflect the dynamic curriculum scheduling of supervision signals.  
This reward integrates multiple sources of supervision, including
(i) \textit{rule-based} verification of final answer correctness, and
(ii) \textit{rubric-based} evaluations that assess intermediate reasoning qualities such as perception, grounding, and logical consistency.  
These rubric signals are derived from a set of evaluation rubrics $\mathcal{R} = \{R_1, \dots, R_k\}$, each targeting a distinct reasoning aspect.  
Our objective is to learn a policy $\pi_\theta$ that maximizes the expected reward:
\begin{equation}
    \max_{\theta} \;
    \mathbb{E}_{x \sim \mathcal{D}, \, y \sim \pi_\theta(\cdot \mid x)}
    \left[ r^{(t)}(y \mid x) \right].
\end{equation}
We optimize this objective using Group Relative Policy Optimization (GRPO)~\cite{shao2024deepseekmath}, a stable policy-gradient method for RLVR (see Appendix~\ref{app:grpo} for details).   
The core challenge is to design $r^{(t)}$ such that it provides adaptive supervision across reasoning skills that differ substantially in difficulty and learnability.
\textbf{Rubric Rewards as Multi-Objective Optimization.}
Rubric-based supervision can be viewed as optimizing multiple skill-wise objectives under a shared policy.
Specifically, each rubric $R_k \in \mathcal{R}$ induces a sub-reward $r_k(y\mid x)$, and the overall rubric reward corresponds to a weighted combination:
\begin{equation}
    r_{\text{rub}}^{(t)}(y\mid x) = \sum_{k=1}^{K} \omega_k^{(t)} \, r_k(y\mid x),
    \quad \text{s.t.} \sum \omega_k^{(t)} = 1.
\end{equation}
A key challenge in this formulation is the heterogeneity of the objectives. The rubrics range from basic checks to complex reasoning steps, implying that their corresponding reward signals vary significantly in density and reliability.
Indiscriminately mixing these diverse signals with static weights risks letting noisy, high-difficulty objectives dominate or interfere with the learning of foundational skills.
Therefore, a time-varying weighting scheme $\omega^{(t)}$ naturally serves as a curriculum over reward components, allowing optimization to prioritize learnable, low-noise objectives first and progressively incorporate harder reasoning criteria.

In the following sections, we detail how $r^{(t)}$ is instantiated via data-driven rubric construction and difficulty stratification (Sec.~\ref{sec:rubric_construction}), and a performance-triggered curriculum scheduling mechanism (Sec.~\ref{sec:reward_scheduling}). The overview of \ours is illustrated in Fig.~\ref{fig:method}.

\subsection{Phase I: Generalized Rubric Construction and Stratification}
\label{sec:rubric_construction}
To construct a robust and discriminative reward system, we design a quantitative, data-driven pipeline that filters and stratifies rubrics based on their empirical behavior. 
In contrast to existing rubric-based methods that generate ad hoc, instance-specific rubrics~\cite{zhou2025breaking,gunjal2025rubrics}, we construct a reusable set of generalized rubrics that remain applicable across diverse reasoning tasks, enabling principled difficulty stratification and curriculum scheduling.

\textbf{Computational Efficiency Analysis.}
We theoretically differentiate the computational overhead of \ours from instance-specific methods~\cite{jia2025autorubric,zhou2025breaking,gunjal2025rubrics}. 
While both paradigms incur a comparable online evaluation cost proportional to the number of training steps
, the critical efficiency gap lies in the \textit{rubric generation} phase.
Let $N$ denote the total number of unique training queries and $C_{gen}$
be the unit costs for generating a rubric set.
Instance-level approaches must synthesize tailored rubrics for every unique input, scaling linearly with the dataset size ($\mathcal{O}(N \times C_{gen})$).
In contrast, \ours generates a generalized rubric pool shared across all data, reducing the generation overhead to a constant $\mathcal{O}(1 \times C_{gen})$.
By eliminating the repetitive LLM calls for per-instance rubric creation, \ours significantly reduces the pre-computation burden without compromising the evaluation density.

\textbf{Candidate Generation \& Rollout.}
We prompt a teacher LLM with comprehensive context, including the task category, relevant images, the input query, and the ground truth answer, and instruct it to generate a diverse set of the most relevant rubric candidates ($\mathcal{R}_{\text{candidates}}$). We then perform rollouts on a randomly sampled subset of training instances ($\mathcal{D}_{\text{sample}}$) of size $N$ using the base model to collect rubric-level evaluation signals.

\textbf{Applicability-Aware Evaluation.}
Unlike standard scalar scoring, we design a specialized Judge mechanism that explicitly decouples \textit{relevance} from \textit{performance}.
For each sample $x_i \in \mathcal{D}_{\text{sample}}$ and rubric candidate $R_j$, the Judge outputs a tuple $(a_{ij}, s_{ij})$, where $a_{ij} \in \{0, 1\}$ indicates whether rubric $R_j$ is applicable to the problem context of $x_i$, and $s_{ij} \in \{0, 1\}$ denotes whether the model output satisfies the rubric, evaluated only when $a_{ij} = 1$.

This explicit decoupling ensures that the computed statistics accurately reflect the rubric's effective coverage and the model's actual proficiency. By preventing non-applicable rubrics from skewing the metrics, this mechanism provides a reliable basis for selecting high-coverage rubrics and stratifying them by difficulty. The detailed evaluation prompt is provided in Appendix~\ref{app:judge_prompts}.

\textbf{Metric-Based Filtering and Stratification.}
Using the assessment statistics, we refine the candidate pool to construct a structured curriculum.
We first compute the \textbf{Applicability Rate ($\eta_j$)} to quantify each rubric's coverage across the dataset: $\eta_j = \frac{1}{N} \sum_{i=1}^N a_{ij}$.
To ensure broad coverage and reduce noise from rarely applicable rubrics, we discard rubrics with insufficient coverage ($\eta_j < \tau_{\text{app}}$).
We provide detailed statistics in Appendix~\ref{app:Applicability and Accuracy} illustrating the high variance in rubric coverage (e.g., as low as 9.7\%), which empirically justifies the necessity of this filtering mechanism to avoid catastrophic gradient noise.
For the remaining rubrics ($\mathcal{R}_{\text{filtered}} \subseteq \mathcal{R}_{\text{candidates}}$), we compute the \textbf{Pass Rate ($p_j$)}, defined as the current model's conditional success rate on applicable instances: $p_j = \frac{\sum_{i=1}^N (a_{ij} \cdot s_{ij})}{\sum_{i=1}^N a_{ij}}$.
This metric serves as an empirical proxy for difficulty, allowing us to stratify rubrics based on their role in the learning process.
We partition them into two distinct levels (see Table~\ref{tab:rubric_reordered}):
\textbf{a) Foundational Rubrics ($\mathcal{R}_{\text{easy}}$)}, characterized by high pass rates, target prerequisite skills to provide stable initial supervision signals;
\textbf{b) Advanced Rubrics ($\mathcal{R}_{\text{hard}}$)}, identified by low pass rates, target complex reasoning gaps that remain underdeveloped in the base model.
This separation enables a curriculum that reinforces basics first, then progressively pivots to challenging reasoning tasks.

\textbf{Statistical Interpretation of Pass Rate as Difficulty Proxy.}
We justify using the pass rate as a principled indicator of optimization difficulty through the lens of gradient estimator stability. For a fixed rubric $R_j$, we model its signal as a Bernoulli variable $r_j \sim \text{Bernoulli}(p_j)$. In policy gradient methods, the reliability of the update is inversely related to the Coefficient of Variation (CV) of the estimator:
\begin{equation}
    CV(r_j) = \frac{\sqrt{Var(r_j)}}{\mathbb{E}[r_j]} = \frac{\sqrt{p_j(1-p_j)}}{p_j} = \sqrt{\frac{1}{p_j} - 1}.
\end{equation}
This derivation reveals a critical insight: as the pass rate $p_j \to 0$, the relative noise diverges ($CV \to \infty$). This implies that rubrics with low pass rates (Advanced Rubrics) provide gradient signals that are dominated by noise, leading to inefficient credit assignment. Conversely, high-pass-rate rubrics (Foundational Rubrics) offer low-CV, reliable signals.
Thus, stratifying rubrics by pass rate is statistically equivalent to stratifying by gradient reliability.
\begin{table*}[t]
    \caption{Performance comparison on Mathematical Reasoning and General benchmarks. The ``Avg.'' column reports the average score across all seven evaluated benchmarks. The best results among open-source reasoning models are highlighted in \textbf{bold}, while the second-best are \underline{underlined}.}
    \centering
    \begin{small}
        \begin{tabular}{l cccc ccc c}
            \toprule
            \multirow{2}{*}{\textbf{Model}} & \multicolumn{4}{c}{\textbf{Mathematical Reasoning}} & \multicolumn{3}{c}{\textbf{General}} & \multirow{2}{*}{\textbf{Avg.}} \\
            \cmidrule(lr){2-5} \cmidrule(lr){6-8}
            & MathVerse & MathVision & MathVista & WeMATH & MMMU & LogicVista & Counting & \\
            \midrule

            \multicolumn{9}{c}{\textbf{Proprietary Models}} \\
            \midrule
            GPT-4o & 50.20 & 30.30 & 63.80 & 68.80 & 69.10 & 45.90 & - & - \\
            Claude-3.5-Sonnet & 57.64 & 46.48 & 67.70 & 73.05 & 68.30 & 43.97 & - & - \\
            \midrule

            \multicolumn{9}{c}{\textbf{Open-Source General-Purpose Models}} \\
            \midrule
            Qwen2.5-VL-7B & 48.98 & 24.18 & 70.20 & 58.52 & 51.00 & 39.26 & 73.50 & 52.23 \\
            Qwen2.5-VL-32B & 57.60 & 38.40 & 74.70 & 69.10 & 57.44 & 49.26 & 85.36 & 61.69 \\
            InternVL2.5-8B & 39.53 & 19.70 & 62.30 & 53.50 & 45.73 & 38.23 & 74.83 & 47.69 \\
            InternVL2.5-38B & 49.40 & 32.20 & 71.84 & 68.61 & 56.98 & 47.21 & 82.77 & 58.43 \\
            \midrule

            \multicolumn{9}{c}{\textbf{Open-Source Multimodal Large Language Models}} \\
            \midrule
            MM-Eureka-7B & 51.09 & 27.70 & 73.00 & 65.34 & 53.78 & \underline{47.87} & 75.50 & 56.33 \\
            OpenVLThinker-7B & 48.37 & 25.90 & 71.38 & 66.63 & 54.29 & 36.24 & 65.00 & 52.54 \\
            Perception-R1-7B & 52.56 & 28.06 & 72.80 & 65.57 & 53.11 & 40.94 & 82.30 & 56.48 \\
            Vision-R1-7B & 53.23 & 27.24 & 70.63 & 64.98 & 43.28 & 42.94 & 83.27 & 55.08 \\
            R1-Onevision-7B & 45.12 & 23.91 & 66.21 & 61.88 & 43.70 & 44.53 & 78.45 & 51.97 \\
            ThinkLite-VL-7B & 51.47 & 27.24 & \underline{73.30} & 65.52 & \underline{55.44} & 42.94 & \textbf{86.50} & \underline{57.49} \\
            VL-Rethinker-7B & \underline{53.86} & \textbf{29.57} & 73.27 & \underline{68.22} & 54.67 & 46.08 & 68.50 & 56.31 \\
            
            \rowcolor{bgblue}
            \textbf{\ours} & \textbf{54.14} & \underline{28.88} & \textbf{74.10} & \textbf{71.49} & \textbf{56.67} & \textbf{49.66} & \underline{85.50} & \textbf{60.06} \\

            
            \textit{$\Delta(\text{Qwen2.5-VL-7B})$} & 
            \textit{\textcolor{customgreen}{$\uparrow$ 5.16}} & 
            \textit{\textcolor{customgreen}{$\uparrow$ 4.70}} & 
            \textit{\textcolor{customgreen}{$\uparrow$ 3.90}} & 
            \textit{\textcolor{customgreen}{$\uparrow$ 12.97}} & 
            \textit{\textcolor{customgreen}{$\uparrow$ 5.67}} & 
            \textit{\textcolor{customgreen}{$\uparrow$ 10.40}} & 
            \textit{\textcolor{customgreen}{$\uparrow$ 12.00}} & 
            \textit{\textcolor{customgreen}{$\uparrow$ 7.83}} \\
            \bottomrule
        \end{tabular}
    \end{small}
    \label{tab:main_results_new}
\end{table*}

\subsection{Phase II: Dynamic Curriculum Learning}
\label{sec:curriculum_learning}

We employ a hybrid reward mechanism that integrates rule-based correctness with the stratified rubrics derived in Phase \hyperref[sec:rubric_construction]{I}. We introduce a stability-aware curriculum that dynamically adjusts the focus from foundational to advanced reasoning.

\textbf{Hybrid Reward Components.}
\label{sec:hybrid_eval}
Our reward system adopts a hybrid evaluation strategy that integrates model-based rubric evaluation with strict rule-based verification, balancing fine-grained rubric-level process supervision with unambiguous outcome correctness.
We employ a strict rule-based verifier to assess the final answer correctness. For each sampled response $y_i$ conditioned on input $x$, the final outcome reward is defined as:
\begin{equation}
    r_{\text{ans}}(y_i \mid x) = \mathbb{I}\!\left(\text{grade}(\hat{y}_i, y^*) = 1\right),
\end{equation}
where $\hat{y}_i$ is the extracted prediction and $y^*$ is the ground truth.

In parallel, we evaluate the reasoning process using the foundational ($\mathcal{R}_{\text{easy}}$) and advanced ($\mathcal{R}_{\text{hard}}$) rubric sets derived in Sec.~\ref{sec:rubric_construction}.
During training, the Judge model evaluates the generated response against all rubrics in these filtered sets. We aggregate the binary satisfaction signals to compute tier-level reasoning scores:
\begin{equation}
\begin{aligned}
    \bar{r}_{\text{easy}}(y_i \mid x) &= \frac{1}{|\mathcal{R}_{\text{easy}}|} \sum_{R \in \mathcal{R}_{\text{easy}}} r(y_i \mid x, R), \\
    \bar{r}_{\text{hard}}(y_i \mid x) &= \frac{1}{|\mathcal{R}_{\text{hard}}|} \sum_{R \in \mathcal{R}_{\text{hard}}} r(y_i \mid x, R),
\end{aligned}
\end{equation}
where $r(y_i \mid x, R) \in \{0, 1\}$ denotes whether the response satisfies rubric $R$. These aggregated scores $\bar{r}_{\text{easy}}$ and $\bar{r}_{\text{hard}}$ serve as the basis of our curriculum scheduling mechanism.

\textbf{Performance-Triggered Curriculum Scheduling.}
\label{sec:reward_scheduling}
We introduce a \textbf{Stability-Aware Curriculum} that regulates the progression from foundational to advanced reasoning supervision.
In contrast to static schedules, \ours activates advanced rubrics only after the model demonstrates stable proficiency on foundational ones.

For a sampled response $y_i$ at training step $t$, we define the curriculum-modulated rubric reward as:
\begin{equation}
\label{eq:rubrics_score}
r_{\text{rub}}^{(t)}(y_i\mid x) = (1 - \lambda_t) \cdot \bar{r}_{\text{easy}}(y_i\mid x) + \lambda_t \cdot \bar{r}_{\text{hard}}(y_i\mid x),
\end{equation}
where the curriculum coefficient $\lambda_t \in [0, \lambda_{\text{max}}]$ controls the difficulty mix between foundational and advanced reasoning rubrics.
Initially, $\lambda_t$ is set to zero and remains unchanged until foundational performance stabilizes.

The curriculum proceeds in three phases:
\paragraph{(1) Stabilization Phase:} We enforce $\lambda_t = 0$. Let $\mu_{\text{easy}}^{(t)} = \mathbb{E}_{(x,y)\sim\mathcal{B}_t}\!\left[\bar{r}_{\text{easy}}(y\mid x)\right]$ denote the batch-averaged foundational rewards at step $t$, and let $W_t=\{\mu_{\text{easy}}^{(t-w+1)},\dots,\mu_{\text{easy}}^{(t)}\}$ be a sliding window of length $w$. The transition is triggered at step $T_{\text{start}}$ only when the model's performance consistently exceeds a proficiency threshold $\tau_{th}$ throughout the entire window:
\begin{equation}
\label{eq:tstart}
    T_{\text{start}} = \min \{ t \mid \forall \mu \in W_t, \mu \geq \tau_{th} \}.
\end{equation}
This strict condition ensures that the model does not progress to advanced reasoning stages due to transient lucky guesses.
    
\paragraph{(2) Curriculum Ramp-up:} Once triggered ($t > T_{\text{start}}$), $\lambda_t$ follows a defined growth function (e.g., Linear or Sigmoid) over a duration $T_{\text{ramp}}$:
\begin{equation}
    \lambda_t = \lambda_{\text{base}} + (\lambda_{\text{max}} - \lambda_{\text{base}}) \cdot \phi\left(\frac{t - T_{\text{start}}}{T_{\text{ramp}}}\right),
\end{equation}
where $\phi(\cdot)$ is the normalized growth function clamped to $[0, 1]$ and $\lambda_{\text{base}}$ denotes the initial curriculum weight.

\paragraph{(3) Advanced Consolidation:} Upon completion of the ramp-up period ($t > T_{\text{start}} + T_{\text{ramp}}$), the curriculum holds the difficulty weight at its peak: $\lambda_t = \lambda_{\text{max}}$.
Finally, we combine the rule-based outcome reward with the curriculum-modulated rubrics reward to obtain the scalar reward used by GRPO:
\begin{equation}
\label{eq:total_reward}
r^{(t)}(y_i\mid x) =
\alpha \cdot r_{\text{ans}}(y_i\mid x)
+
(1 - \alpha) \cdot r_{\text{rub}}^{(t)}(y_i\mid x).
\end{equation}
Here,  $r^{(t)}(y_i\mid x)$ is the scalar reward used in GRPO advantage estimation. We treat $\alpha \in [0,1]$ as a fixed hyperparameter that controls the trade-off between outcome correctness and rubric-based process supervision.

\textbf{Analysis.}
While traditional curriculum learning operates by reshaping the input distribution, \ours instead modulates the density of evaluative signals over the output space. We posit a hierarchical dependency among rubrics, where satisfying advanced (hard) rubrics presupposes competence in foundational (easy) ones. We further provide a theoretical justification for this design in Appendix~\ref{app:gradient_variance}, demonstrating that the proposed schedule reduces the contribution of unreliable and high-noise gradient components induced by sparse advanced rewards, thereby stabilizing early-stage optimization. Prioritizing easy rubrics early in training therefore performs an implicit search-space pruning, restricting optimization to regions of the policy space where advanced rubric signals become attainable. This design reduces gradient interference from currently unachievable objectives, alleviates cold-start instability, and yields more stable and efficient optimization throughout training.

\section{Experiments}

\subsection{Experiment Setup}
\textbf{Datasets \& Models.}
In our experiments, we utilize the ViRL-39K dataset~\citep{wang2025vl} for model training. 
ViRL-39K is a large-scale, high-quality dataset specifically curated for vision-language reinforcement learning (RL). It comprises approximately 39,000 verifiable question-answering pairs that cover a wide range of complex scenarios, including STEM, spatial reasoning, and multi-disciplinary chart analysis. 
Specifically, we initialize our training from Qwen2.5-VL-7B-Instruct ~\cite{bai2025qwen2} as the base model, leveraging its advanced multi-modal perception and robust instruction-following capabilities to facilitate further reasoning-oriented optimization.

\textbf{Evaluation.}
We evaluate \ours on widely used visual reasoning benchmarks covering multimodal mathematical reasoning and general visual reasoning.
For multimodal mathematical reasoning, we use MathVista~\citep{lu2023mathvista}, MathVerse~\citep{zhang2024mathverse}, MATH-Vision~\citep{wang2024measuring}, and WeMATH~\citep{qiao2025we}. 
For general visual reasoning, we employ LogicVista~\citep{xiao2024logicvista}, Super-CLEVR Counting~\citep{li2023super}, and MMMU~\citep{yue2024mmmu} to assess logical deduction, compositional counting and perception, and multi-disciplinary knowledge, respectively.

\textbf{Baselines.}
We compare our model with several strong MLLMs, categorized into three groups: 
(1) \textit{Proprietary models}, including GPT-4o~\cite{hurst2024gpt} and Claude-3.5-Sonnet~\cite{anthropic2024}; 
(2) \textit{Open-source general-purpose models}, such as Qwen2.5-VL-7B-Instruct, Qwen2.5-VL-32B-Instruct~\cite{bai2025qwen2}, InternVL2.5-8B
and InternVL2.5-38B~\cite{chen2024expanding}; and 
(3) \textit{Open-source reasoning-focused models}, including MM-Eureka-7B~\cite{meng2025mm}, OpenVLThinker-7B~\cite{deng2025openvlthinker}, Perception-R1-7B~\cite{xiao2025advancing}, Vision-R1-7B~\cite{huang2025vision}, R1-Onevision-7B~\cite{yang2025r1}, ThinkLite-VL-7B~\cite{wang2025sota}, and VL-Rethinker-7B~\cite{wang2025vl}.

\textbf{Configuration.}
For data-driven candidate generation, we utilize Gemini 3 Pro~\cite{google2025gemini3} as the teacher model. Through few-shot prompting, we generate 20 rubric candidates. Subsequently, we conduct a rollout on $N=2,000$ samples, retaining 6 core rubrics after filtering with an applicability threshold of $0.99$.
In the reinforcement learning phase, we deploy Qwen3-VL-235B-A22B-Instruct~\cite{qwen3technicalreport} as the reward judge. The detailed prompts guiding the judge model's scoring process are provided in Appendix~\ref{app:reward_prompts}. To implement the proposed \textit{Stability-Aware Curriculum}, we configure the sliding window size $K=20$ and the proficiency threshold $\tau_{th}=0.9$. The reward balancing coefficient is set to $\alpha=0.7$ to prioritize factual accuracy. All experiments are conducted on NVIDIA H200 GPUs using the \texttt{verl} framework~\cite{sheng2025hybridflow}. Comprehensive hyperparameter details are provided in Appendix~\ref{app:configuration}.

\begin{figure*}[!t]
  \centering
  \includegraphics[width=\linewidth]{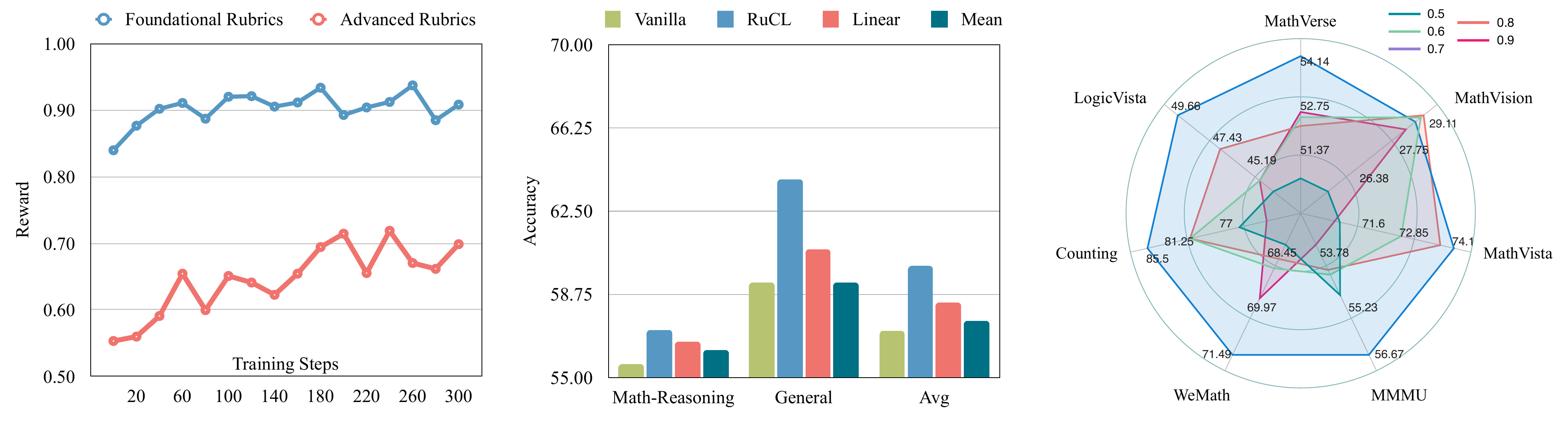}
  \caption{\textbf{Left:} Training dynamics of Foundational (blue) and Advanced (red) rubric rewards. \textbf{Middle:} Ablation study results on rubric aggregation and scheduling strategies. \textbf{Right:} Sensitivity analysis of the reward balancing hyperparameter.}
  \label{fig:result}
\end{figure*}

\subsection{Main Results}
\label{sec:results}

\paragraph{Mathematical Reasoning Performance.}
As shown in Table~\ref{tab:main_results_new}, \ours demonstrates superior performance, outperforming the baseline Qwen2.5-VL-7B across all mathematical benchmarks. This significant improvement, driven by the integration of our fine-grained rubric-based reward modeling and curriculum learning strategy, validates the efficacy of prioritizing simple rubrics in early training stages before transitioning to harder reasoning constraints. Specifically, on the challenging WeMATH and MathVerse datasets, our model improves by 12.97\% (from 58.52\% to 71.49\%) and 5.16\% (from 48.98\% to 54.14\%), respectively. Furthermore, when compared with other leading open-source reasoning models such as ThinkLite-VL-7B and VL-Rethinker-7B, \ours achieves the highest average score of 60.06\% across all seven tasks, highlighting its robust reasoning capabilities.

\textbf{Generalization to General and Logical Benchmarks.}
Extending beyond mathematics, our model achieves competitive results across broader reasoning tasks. As shown in the \textit{General} section of Table~\ref{tab:main_results_new}, \ours exhibits remarkable generalization. On the LogicVista benchmark, which requires complex logical deduction, our model achieves a 10.40\% improvement over the baseline (from 39.26\% to 49.66\%), surpassing all other open-source 7B competitors. Similarly, we observe substantial gains on the comprehensive MMMU (+5.67\%) and Counting (+12.00\%) benchmarks, with the latter highlighting enhanced fine-grained visual perception (85.50\%). These results indicate that combining intermediate rubric rewards with final outcome supervision effectively enhances the model's fundamental reasoning robustness rather than merely overfitting to mathematical domains. Notably, despite its compact scale, our model significantly narrows the performance gap with top-tier proprietary models.


\textbf{Training Dynamics and Curriculum Efficacy.}
To validate the efficacy of \ours, we analyze the evolution of reward trajectories throughout the training process, as shown in Figure~\ref{fig:result}. Initially, the curriculum prioritizes foundational rubrics, leading to the rapid mastery of prerequisite skills such as visual presence and entity extraction. As the mechanism detects stable proficiency (scores stabilizing $>0.9$) and progressively introduces advanced reasoning constraints, the model exhibits steady improvement in higher-order tasks while maintaining robust performance on foundational metrics. This demonstrates that \ours fosters complex reasoning while preserving foundational visual perception and instruction-following skills. Furthermore, qualitative case studies in Appendix~\ref{app:case study} provide concrete evidence of \ours's capability to mitigate reward hacking. We show that our rubric-based judge effectively penalizes spurious reasoning chains that serendipitously arrive at the correct answer—instances that typically escape detection in outcome-only supervision—thereby enforcing genuine logical consistency.


\subsection{Ablation Study}
\label{sec:ablation}

In this section, we conduct ablation studies to validate the contributions of our key design choices. We focus on two key components: the rubric aggregation mechanism and the sensitivity to the reward balancing hyperparameter $\alpha$.

\textbf{Impact of Rubric Aggregation and Scheduling.}
To assess the contribution of rubric aggregation and curriculum scheduling, we compare our method (\textbf{Sigmoid Stratification}) with the following baselines, keeping the GRPO backbone and training data fixed: \textbf{(1) Vanilla GRPO:} Trains using solely the rule-based outcome reward $r_{\text{ans}}$, ignoring all reasoning rubrics. \textbf{(2) Uniform Averaging:} Aggregates all filtered rubrics into a single unweighted average score, discarding difficulty stratification and curriculum scheduling. \textbf{(3) \ours (Sigmoid Stratification):} Adopts the proposed stratified rubrics ($\mathcal{R}_{\text{easy}}, \mathcal{R}_{\text{hard}}$) with the stability-aware sigmoid schedule for $\lambda_t$. \textbf{(4) Linear Stratification:} Replaces the sigmoid growth function with a simple linear ramp for $\lambda_t$ to evaluate the impact of schedule shape.



\begin{table}[t]
    \caption{Ablation results on General benchmarks.}
    \centering
    \resizebox{\linewidth}{!}{
        \begin{tabular}{l cccc}
            \toprule
            \textbf{Method} & \textbf{MMMU} & \textbf{LogicVista} & \textbf{Counting} & \textbf{Avg.} \\
            \midrule
            Vanilla GRPO & 54.89 & 46.53 & 76.00 & 59.14 \\
            Uniform Averaging & 55.44 & \textbf{50.11} & 77.00 & 59.29 \\
            Linear Stratification & 55.44 & 47.43 & 79.50 & 60.79 \\
            \textbf{\ours} & \textbf{56.67} & 49.66 & \textbf{85.50} & \textbf{63.94} \\
            \bottomrule
        \end{tabular}
    }
    \vspace{-0.3cm}
    \label{tab:ablation_general}
\end{table}

Figure~\ref{fig:result} highlights the aggregate trend: \textit{Vanilla GRPO} (57.13\%) is surpassed by \textit{Uniform Averaging} (57.56\%) due to process supervision, while \textit{Linear Stratification} (58.41\%) yields further gains by distinguishing difficulty. As shown in Table~\ref{tab:ablation_general}, \ours significantly outperforms the Linear strategy, particularly on perception-heavy tasks like Counting (85.50\% vs. 79.50\%) and logic-intensive tasks like LogicVista (49.66\% vs. 47.43\%). This advantage stems from the Sigmoid schedule's ability to reach maximum difficulty saturation earlier than the linear ramp. By completing the transition phase faster, Sigmoid affords the model a longer stable period to converge under the full weight of hard constraints, whereas the Linear approach keeps the reward signal in a continuous state of flux.

\textbf{Sensitivity to Reward Balancing Hyperparameter 
\texorpdfstring{$\alpha$}{alpha}.}
\label{sec:sensitivity}
We further investigate the system's sensitivity to the hyperparameter $\alpha \in [0.5, 0.9]$, which governs the trade-off between outcome correctness and rubric-based fine-grained supervision. As visualized in the radar chart (Figure~\ref{fig:result}), the overall performance achieves its optimum at $\alpha = 0.7$. At this optimal setting, the model demonstrates robust dominance across diverse tasks, achieving peak scores of 71.49\% on WeMath and 85.50\% on Counting. Deviating from this balance proves detrimental: lowering $\alpha$ to 0.5 (where fine-grained rubrics dominate) causes performance drops (e.g., Counting falls to 79.00\%), likely because excessive auxiliary constraints distract from the primary objective of solution correctness. Conversely, increasing $\alpha$ to 0.9 diminishes the benefit of our fine-grained supervision, causing the system to degenerate towards a sparse-reward regime where complex reasoning capability degrades significantly (e.g., WeMath drops to 53.78\%). Thus, a configuration of $\alpha = 0.7$ strikes the most effective balance, integrating precise intermediate guidance without overshadowing the ultimate goal of accurate problem-solving.

\begin{table}[t]
    \caption{Sensitivity analysis of sliding window size $w$ in curriculum triggering.}
    \centering
        \begin{tabular}{l c c c}
            \toprule
            \textbf{Window Size $w$} & \textbf{Mathematical} & \textbf{General} & \textbf{Avg.} \\
            \midrule
            $w=10$ & 55.57 & 60.8 &  57.81 \\
            $w=20$ & \textbf{57.15} & \textbf{63.94}  & \textbf{60.06} \\
            $w=30$ & 56.64 & 63.74 & 59.67 \\
            \bottomrule
        \end{tabular}
    \label{tab:ablation_window}
\end{table}

\textbf{Sensitivity to Sliding Window Size $w$.}
We study the sensitivity of the stability-aware trigger to the sliding window size $w$ in Eq.~\ref{eq:tstart}.
We vary $w \in \{10, 20, 30\}$ while keeping all other hyperparameters fixed.
As shown in Table~\ref{tab:ablation_window}, $w=20$ achieves the best overall performance, while $w=30$ performs comparably with only a marginal gap.
In contrast, $w=10$ consistently underperforms, indicating that a shorter window is more susceptible to transient fluctuations and may trigger the curriculum transition prematurely.
Overall, these observations suggest that our curriculum mechanism is robust to moderate changes in $w$, and we adopt $w=20$ as the default setting in all experiments.





\section{Conclusion}

We propose \textbf{Stratified Rubric-based Curriculum Learning (\ours)}, a framework that reframes curriculum learning from data selection to reward design. By stratifying evaluation rubrics into foundational and advanced categories, \ours aligns reward signals with the model's evolving capabilities. Integrated with GRPO, this approach effectively mitigates reward hacking and training instability. Experiments across seven benchmarks demonstrate that \ours significantly outperforms the base model and establishes a new state-of-the-art among 7B-scale reasoning models. Future work will explore online rubric construction and scaling to larger architectures.

\section*{Impact Statement}
This paper introduces Stratified Rubric-based Curriculum Learning (RuCL), a framework that enhances the reasoning capabilities of Multimodal Large Language Models (MLLMs) by shifting curriculum focus from data selection to reward design. \ours guides models to master foundational perception before progressing to advanced deduction, fostering the development of reliable models that prioritize intermediate reasoning integrity. \ours utilizes widely recognized, publicly available datasets for training and evaluation, strictly adhering to their licenses and usage policies without intentionally introducing private, personally identifiable information (PII) or offensive content, ensuring that our advancements in multimodal intelligence are built upon transparent and reproducible foundations.


\bibliography{example_paper}

@inproceedings{yue2024mmmu,
  title={Mmmu: A massive multi-discipline multimodal understanding and reasoning benchmark for expert agi},
  author={Yue, Xiang and Ni, Yuansheng and Zhang, Kai and Zheng, Tianyu and Liu, Ruoqi and Zhang, Ge and Stevens, Samuel and Jiang, Dongfu and Ren, Weiming and Sun, Yuxuan and others},
  booktitle={Proceedings of the IEEE/CVF Conference on Computer Vision and Pattern Recognition},
  pages={9556--9567},
  year={2024}
}

@inproceedings{bengio2009curriculum,
  title={Curriculum learning},
  author={Bengio, Yoshua and Louradour, J{\'e}r{\^o}me and Collobert, Ronan and Weston, Jason},
  booktitle={Proceedings of the 26th annual international conference on machine learning},
  pages={41--48},
  year={2009}
}

@article{yuan2025vl,
  title={Vl-cogito: Progressive curriculum reinforcement learning for advanced multimodal reasoning},
  author={Yuan, Ruifeng and Xiao, Chenghao and Leng, Sicong and Wang, Jianyu and Li, Long and Xu, Weiwen and Chan, Hou Pong and Zhao, Deli and Xu, Tingyang and Wei, Zhongyu and others},
  journal={arXiv preprint arXiv:2507.22607},
  year={2025}
}

@article{song2025fastcurl,
  title={FastCuRL: Curriculum Reinforcement Learning with Stage-wise Context Scaling for Efficient Training R1-like Reasoning Models},
  author={Song, Mingyang and Zheng, Mao and Li, Zheng and Yang, Wenjie and Luo, Xuan and Pan, Yue and Zhang, Feng},
  journal={arXiv preprint arXiv:2503.17287},
  year={2025}
}

@article{parashar2025curriculum,
  title={Curriculum Reinforcement Learning from Easy to Hard Tasks Improves LLM Reasoning},
  author={Parashar, Shubham and Gui, Shurui and Li, Xiner and Ling, Hongyi and Vemuri, Sushil and Olson, Blake and Li, Eric and Zhang, Yu and Caverlee, James and Kalathil, Dileep and others},
  journal={arXiv preprint arXiv:2506.06632},
  year={2025}
}

@article{schulman2017proximal,
  title={Proximal policy optimization algorithms},
  author={Schulman, John and Wolski, Filip and Dhariwal, Prafulla and Radford, Alec and Klimov, Oleg},
  journal={arXiv preprint arXiv:1707.06347},
  year={2017}
}

@article{chen2025self,
  title={Self-Evolving Curriculum for LLM Reasoning},
  author={Chen, Xiaoyin and Lu, Jiarui and Kim, Minsu and Zhang, Dinghuai and Tang, Jian and Pich{\'e}, Alexandre and Gontier, Nicolas and Bengio, Yoshua and Kamalloo, Ehsan},
  journal={arXiv preprint arXiv:2505.14970},
  year={2025}
}

@article{shi2025efficient,
  title={Efficient reinforcement finetuning via adaptive curriculum learning},
  author={Shi, Taiwei and Wu, Yiyang and Song, Linxin and Zhou, Tianyi and Zhao, Jieyu},
  journal={arXiv preprint arXiv:2504.05520},
  year={2025}
}

@article{team2025kwai,
  title={Kwai Keye-VL Technical Report},
  author={Team, Kwai Keye and Yang, Biao and Wen, Bin and Liu, Changyi and Chu, Chenglong and Song, Chengru and Rao, Chongling and Yi, Chuan and Li, Da and Zang, Dunju and others},
  journal={arXiv preprint arXiv:2507.01949},
  year={2025}
}

@article{lu2023mathvista,
  title={Mathvista: Evaluating mathematical reasoning of foundation models in visual contexts},
  author={Lu, Pan and Bansal, Hritik and Xia, Tony and Liu, Jiacheng and Li, Chunyuan and Hajishirzi, Hannaneh and Cheng, Hao and Chang, Kai-Wei and Galley, Michel and Gao, Jianfeng},
  journal={arXiv preprint arXiv:2310.02255},
  year={2023}
}

@article{hurst2024gpt,
  title={Gpt-4o system card},
  author={Hurst, Aaron and Lerer, Adam and Goucher, Adam P and Perelman, Adam and Ramesh, Aditya and Clark, Aidan and Ostrow, AJ and Welihinda, Akila and Hayes, Alan and Radford, Alec and others},
  journal={arXiv preprint arXiv:2410.21276},
  year={2024}
}

@article{chen2024expanding,
  title={Expanding performance boundaries of open-source multimodal models with model, data, and test-time scaling},
  author={Chen, Zhe and Wang, Weiyun and Cao, Yue and Liu, Yangzhou and Gao, Zhangwei and Cui, Erfei and Zhu, Jinguo and Ye, Shenglong and Tian, Hao and Liu, Zhaoyang and others},
  journal={arXiv preprint arXiv:2412.05271},
  year={2024}
}

@article{team2025kimi,
  title={Kimi-vl technical report},
  author={Team, Kimi and Du, Angang and Yin, Bohong and Xing, Bowei and Qu, Bowen and Wang, Bowen and Chen, Cheng and Zhang, Chenlin and Du, Chenzhuang and Wei, Chu and others},
  journal={arXiv preprint arXiv:2504.07491},
  year={2025}
}

@article{meng2025mm,
  title={Mm-eureka: Exploring the frontiers of multimodal reasoning with rule-based reinforcement learning},
  author={Meng, Fanqing and Du, Lingxiao and Liu, Zongkai and Zhou, Zhixiang and Lu, Quanfeng and Fu, Daocheng and Han, Tiancheng and Shi, Botian and Wang, Wenhai and He, Junjun and others},
  journal={arXiv preprint arXiv:2503.07365},
  year={2025}
}

@article{xiao2025advancing,
  title={Advancing Multimodal Reasoning Capabilities of Multimodal Large Language Models via Visual Perception Reward},
  author={Xiao, Tong and Xu, Xin and Huang, Zhenya and Gao, Hongyu and Liu, Quan and Liu, Qi and Chen, Enhong},
  journal={arXiv preprint arXiv:2506.07218},
  year={2025}
}

@article{huang2025vision,
  title={Vision-r1: Incentivizing reasoning capability in multimodal large language models},
  author={Huang, Wenxuan and Jia, Bohan and Zhai, Zijie and Cao, Shaosheng and Ye, Zheyu and Zhao, Fei and Xu, Zhe and Hu, Yao and Lin, Shaohui},
  journal={arXiv preprint arXiv:2503.06749},
  year={2025}
}

@article{wang2025sota,
  title={Sota with less: Mcts-guided sample selection for data-efficient visual reasoning self-improvement},
  author={Wang, Xiyao and Yang, Zhengyuan and Feng, Chao and Lu, Hongjin and Li, Linjie and Lin, Chung-Ching and Lin, Kevin and Huang, Furong and Wang, Lijuan},
  journal={arXiv preprint arXiv:2504.07934},
  year={2025}
}

@article{bai2025qwen2,
  title={Qwen2. 5-vl technical report},
  author={Bai, Shuai and Chen, Keqin and Liu, Xuejing and Wang, Jialin and Ge, Wenbin and Song, Sibo and Dang, Kai and Wang, Peng and Wang, Shijie and Tang, Jun and others},
  journal={arXiv preprint arXiv:2502.13923},
  year={2025}
}

@article{deng2025openvlthinker,
  title={Openvlthinker: An early exploration to complex vision-language reasoning via iterative self-improvement},
  author={Deng, Yihe and Bansal, Hritik and Yin, Fan and Peng, Nanyun and Wang, Wei and Chang, Kai-Wei},
  journal={arXiv preprint arXiv:2503.17352},
  year={2025}
}

@inproceedings{zhang2024mathverse,
  title={Mathverse: Does your multi-modal llm truly see the diagrams in visual math problems?},
  author={Zhang, Renrui and Jiang, Dongzhi and Zhang, Yichi and Lin, Haokun and Guo, Ziyu and Qiu, Pengshuo and Zhou, Aojun and Lu, Pan and Chang, Kai-Wei and Qiao, Yu and others},
  booktitle={European Conference on Computer Vision},
  pages={169--186},
  year={2024},
  organization={Springer}
}

@article{wang2024measuring,
  title={Measuring multimodal mathematical reasoning with math-vision dataset},
  author={Wang, Ke and Pan, Junting and Shi, Weikang and Lu, Zimu and Ren, Houxing and Zhou, Aojun and Zhan, Mingjie and Li, Hongsheng},
  journal={Advances in Neural Information Processing Systems},
  volume={37},
  pages={95095--95169},
  year={2024}
}

@inproceedings{qiao2025we,
  title={We-math: Does your large multimodal model achieve human-like mathematical reasoning?},
  author={Qiao, Runqi and Tan, Qiuna and Dong, Guanting and MinhuiWu, MinhuiWu and Sun, Chong and Song, Xiaoshuai and Wang, Jiapeng and GongQue, Zhuoma and Lei, Shanglin and Zhang, Yifan and others},
  booktitle={Proceedings of the 63rd Annual Meeting of the Association for Computational Linguistics (Volume 1: Long Papers)},
  pages={20023--20070},
  year={2025}
}

@article{xu2025evade,
  title={EVADE: Multimodal Benchmark for Evasive Content Detection in E-Commerce Applications},
  author={Xu, Ancheng and Yang, Zhihao and Li, Jingpeng and Yuan, Guanghu and Chen, Longze and Yan, Liang and Zhou, Jiehui and Qin, Zhen and Chang, Hengyun and Alinejad-Rokny, Hamid and others},
  journal={arXiv preprint arXiv:2505.17654},
  year={2025}
}

@article{shao2024deepseekmath,
  title={Deepseekmath: Pushing the limits of mathematical reasoning in open language models},
  author={Shao, Zhihong and Wang, Peiyi and Zhu, Qihao and Xu, Runxin and Song, Junxiao and Bi, Xiao and Zhang, Haowei and Zhang, Mingchuan and Li, YK and Wu, Yang and others},
  journal={arXiv preprint arXiv:2402.03300},
  year={2024}
}

@article{mahdaoui2025automated,
  title={Automated Grading Method of Python Code Submissions Using Large Language Models and Machine Learning},
  author={Mahdaoui, Mariam and Nouh, Said and El Kasmi Alaoui, My Seddiq and Kandali, Khalid},
  journal={Information},
  volume={16},
  number={8},
  pages={674},
  year={2025},
  publisher={MDPI}
}

@article{viswanathan2025checklists,
  title={Checklists are better than reward models for aligning language models},
  author={Viswanathan, Vijay and Sun, Yanchao and Ma, Shuang and Kong, Xiang and Cao, Meng and Neubig, Graham and Wu, Tongshuang},
  journal={arXiv preprint arXiv:2507.18624},
  year={2025}
}

@article{gunjal2025rubrics,
  title={Rubrics as rewards: Reinforcement learning beyond verifiable domains},
  author={Gunjal, Anisha and Wang, Anthony and Lau, Elaine and Nath, Vaskar and He, Yunzhong and Liu, Bing and Hendryx, Sean},
  journal={arXiv preprint arXiv:2507.17746},
  year={2025}
}

@article{ma2025s,
  title   = {S$^{2}$R: Teaching LLMs to Self-verify and Self-correct via Reinforcement Learning},
  author  = {Ma, Ruotian and Wang, Peisong and Liu, Cheng and Liu, Xingyan and Chen, Jiaqi and Zhang, Bang and Zhou, Xin and Du, Nan and Li, Jia},
  journal = {arXiv preprint arXiv:2502.12853},
  year    = {2025}
}

@article{chu2025gpg,
  title   = {Gpg: A simple and strong reinforcement learning baseline for model reasoning},
  author  = {Chu, Xiangxiang and Huang, Hailang and Zhang, Xiao and Wei, Fei and Wang, Yong},
  journal = {arXiv preprint arXiv:2504.02546},
  year    = {2025}
}

@inproceedings{pathak2025rubric,
  title={Rubric is all you need: Improving llm-based code evaluation with question-specific rubrics},
  author={Pathak, Aditya and Gandhi, Rachit and Uttam, Vaibhav and Ramamoorthy, Arnav and Ghosh, Pratyush and Jindal, Aaryan Raj and Verma, Shreyash and Mittal, Aditya and Ased, Aashna and Khatri, Chirag and others},
  booktitle={Proceedings of the 2025 ACM Conference on International Computing Education Research V. 1},
  pages={181--195},
  year={2025}
}

@misc{arora2025healthbenchevaluatinglargelanguage,
      title={HealthBench: Evaluating Large Language Models Towards Improved Human Health}, 
      author={Rahul K. Arora and Jason Wei and Rebecca Soskin Hicks and Preston Bowman and Joaquin Quiñonero-Candela and Foivos Tsimpourlas and Michael Sharman and Meghan Shah and Andrea Vallone and Alex Beutel and Johannes Heidecke and Karan Singhal},
      year={2025},
      eprint={2505.08775},
      archivePrefix={arXiv},
      primaryClass={cs.CL},
      url={https://arxiv.org/abs/2505.08775}, 
}

@article{galvan2025rubrik,
  title={Rubrik's Cube: Testing a New Rubric for Evaluating Explanations on the CUBE dataset},
  author={Galvan-Sosa, Diana and Gaudeau, Gabrielle and Kavumba, Pride and Li, Yunmeng and Yuan, Zheng and Sakaguchi, Keisuke and Buttery, Paula and others},
  journal={arXiv preprint arXiv:2503.23899},
  year={2025}
}

@inproceedings{fan2024sedareval,
  title={Sedareval: Automated evaluation using self-adaptive rubrics},
  author={Fan, Zhiyuan and Wang, Weinong and Zhang, Debing and others},
  booktitle={Findings of the Association for Computational Linguistics: EMNLP 2024},
  pages={16916--16930},
  year={2024}
}

@article{winata2025datasheets,
  title={Datasheets Aren't Enough: DataRubrics for Automated Quality Metrics and Accountability},
  author={Winata, Genta Indra and Anugraha, David and Liu, Emmy and Aji, Alham Fikri and Hung, Shou-Yi and Parashar, Aditya and Irawan, Patrick Amadeus and Zhang, Ruochen and Yong, Zheng-Xin and Cruz, Jan Christian Blaise and others},
  journal={arXiv preprint arXiv:2506.01789},
  year={2025}
}

@inproceedings{zhang2025improve,
  title={Improve vision language model chain-of-thought reasoning},
  author={Zhang, Ruohong and Zhang, Bowen and Li, Yanghao and Zhang, Haotian and Sun, Zhiqing and Gan, Zhe and Yang, Yinfei and Pang, Ruoming and Yang, Yiming},
  booktitle={Proceedings of the 63rd Annual Meeting of the Association for Computational Linguistics (Volume 1: Long Papers)},
  pages={1631--1662},
  year={2025}
}

@article{wang2024enhancing,
  title={Enhancing the reasoning ability of multimodal large language models via mixed preference optimization},
  author={Wang, Weiyun and Chen, Zhe and Wang, Wenhai and Cao, Yue and Liu, Yangzhou and Gao, Zhangwei and Zhu, Jinguo and Zhu, Xizhou and Lu, Lewei and Qiao, Yu and others},
  journal={arXiv preprint arXiv:2411.10442},
  year={2024}
}

@article{rafailov2023direct,
  title={Direct preference optimization: Your language model is secretly a reward model},
  author={Rafailov, Rafael and Sharma, Archit and Mitchell, Eric and Manning, Christopher D and Ermon, Stefano and Finn, Chelsea},
  journal={Advances in neural information processing systems},
  volume={36},
  pages={53728--53741},
  year={2023}
}

@misc{deepscaler2025,
  title={DeepScaleR: Surpassing O1-Preview with a 1.5B Model by Scaling RL},
  author={Michael Luo and Sijun Tan and Justin Wong and Xiaoxiang Shi and William Y. Tang and Manan Roongta and Colin Cai and Jeffrey Luo and Li Erran Li and Raluca Ada Popa and Ion Stoica},
  year={2025},
  note={Notion Blog}
}

@misc{google2025gemini3,
  author = {{Google DeepMind}},
  title = {Gemini 3 Pro},
  howpublished = {\url{https://deepmind.google/technologies/gemini/}},
  year = {2025},
  note = {Accessed: 2025-12-23}
}

@misc{qwen3technicalreport,
      title={Qwen3 Technical Report}, 
      author={Qwen Team},
      year={2025},
      eprint={2505.09388},
      archivePrefix={arXiv},
      primaryClass={cs.CL},
      url={https://arxiv.org/abs/2505.09388}, 
}

@article{xiao2024logicvista,
  title={LogicVista: Multimodal LLM Logical Reasoning Benchmark in Visual Contexts},
  author={Xiao, Yijia and Hu, Yue and Tan, Jiachen and Hu, Pan and Guo, Xiaojun and others},
  journal={arXiv preprint arXiv:2407.04973},
  year={2024}
}

@article{yang2025r1,
  title={R1-onevision: Advancing generalized multimodal reasoning through cross-modal formalization},
  author={Yang, Yi and He, Xiaoxuan and Pan, Hongkun and Jiang, Xiyan and Deng, Yan and Yang, Xingtao and Lu, Haoyu and Yin, Dacheng and Rao, Fengyun and Zhu, Minfeng and others},
  journal={arXiv preprint arXiv:2503.10615},
  year={2025}
}

@article{sheng2024hybridflow,
  title   = {HybridFlow: A Flexible and Efficient RLHF Framework},
  author  = {Guangming Sheng and Chi Zhang and Zilingfeng Ye and Xibin Wu and Wang Zhang and Ru Zhang and Yanghua Peng and Haibin Lin and Chuan Wu},
  year    = {2024},
  journal = {arXiv preprint arXiv: 2409.19256}
}

@inproceedings{sheng2025hybridflow,
  title={Hybridflow: A flexible and efficient rlhf framework},
  author={Sheng, Guangming and Zhang, Chi and Ye, Zilingfeng and Wu, Xibin and Zhang, Wang and Zhang, Ru and Peng, Yanghua and Lin, Haibin and Wu, Chuan},
  booktitle={Proceedings of the Twentieth European Conference on Computer Systems},
  pages={1279--1297},
  year={2025}
}

@misc{anthropic2024,
  author = {{Anthropic}},
  title = {Claude 3.5 Sonnet Model Card Addendum},
  year = {2024},
  howpublished = {\url{https://www.anthropic.com/news/claude-3-5-sonnet}},
  note = {Accessed: 2025-12-23}
}

@article{jia2025autorubric,
  title={AutoRubric-R1V: Rubric-Based Generative Rewards for Faithful Multimodal Reasoning},
  author={Jia, Mengzhao and Zhang, Zhihan and Cases, Ignacio and Liu, Zheyuan and Jiang, Meng and Qi, Peng},
  journal={arXiv preprint arXiv:2510.14738},
  year={2025}
}

@inproceedings{li2023super,
  title={Super-CLEVR: A Virtual Benchmark to Diagnose Domain Robustness in Visual Reasoning},
  author={Li, Zhuowan and Wang, Xingrui and Stengel-Eskin, Elias and Kortylewski, Adam and Ma, Wufei and Van Durme, Benjamin and Yuille, Alan L},
  booktitle={Proceedings of the IEEE/CVF Conference on Computer Vision and Pattern Recognition},
  pages={14963--14973},
  year={2023}
}

@article{zhou2025breaking,
  title={Breaking the exploration bottleneck: Rubric-scaffolded reinforcement learning for general llm reasoning},
  author={Zhou, Yang and Li, Sunzhu and Liu, Shunyu and Fang, Wenkai and Zhang, Kongcheng and Zhao, Jiale and Yang, Jingwen and Zhou, Yihe and Lv, Jianwei and Zheng, Tongya and others},
  journal={arXiv preprint arXiv:2508.16949},
  year={2025}
}

@inproceedings{wen2025light,
  title={Light-r1: Curriculum sft, dpo and rl for long cot from scratch and beyond},
  author={Wen, Liang and Cai, Yunke and Xiao, Fenrui and He, Xin and An, Qi and Duan, Zhenyu and Du, Yimin and Liu, Junchen and Tanglifu, Tanglifu and Lv, Xiaowei and others},
  booktitle={Proceedings of the 63rd Annual Meeting of the Association for Computational Linguistics (Volume 6: Industry Track)},
  pages={318--327},
  year={2025}
}

@article{wang2025vl,
  title={Vl-rethinker: Incentivizing self-reflection of vision-language models with reinforcement learning},
  author={Wang, Haozhe and Qu, Chao and Huang, Zuming and Chu, Wei and Lin, Fangzhen and Chen, Wenhu},
  journal={arXiv preprint arXiv:2504.08837},
  year={2025}
}

@article{wan2025srpo,
  title={Srpo: Enhancing multimodal llm reasoning via reflection-aware reinforcement learning},
  author={Wan, Zhongwei and Dou, Zhihao and Liu, Che and Zhang, Yu and Cui, Dongfei and Zhao, Qinjian and Shen, Hui and Xiong, Jing and Xin, Yi and Jiang, Yifan and others},
  journal={arXiv preprint arXiv:2506.01713},
  year={2025}
}

@article{zhan2025gthinker,
  title={GThinker: Towards General Multimodal Reasoning via Cue-Guided Rethinking},
  author={Zhan, Yufei and Wu, Ziheng and Zhu, Yousong and Xue, Rongkun and Luo, Ruipu and Chen, Zhenghao and Zhang, Can and Li, Yifan and He, Zhentao and Yang, Zheming and others},
  journal={arXiv preprint arXiv:2506.01078},
  year={2025}
}

@article{yao2024mulberry,
  title={Mulberry: Empowering mllm with o1-like reasoning and reflection via collective monte carlo tree search},
  author={Yao, Huanjin and Huang, Jiaxing and Wu, Wenhao and Zhang, Jingyi and Wang, Yibo and Liu, Shunyu and Wang, Yingjie and Song, Yuxin and Feng, Haocheng and Shen, Li and others},
  journal={arXiv preprint arXiv:2412.18319},
  year={2024}
}

@article{liu2025othink,
  title={OThink-MR1: Stimulating multimodal generalized reasoning capabilities via dynamic reinforcement learning},
  author={Liu, Zhiyuan and Zhang, Yuting and Liu, Feng and Zhang, Changwang and Sun, Ying and Wang, Jun},
  journal={arXiv preprint arXiv:2503.16081},
  year={2025}
}

@article{peng2025skywork,
  title={Skywork r1v: Pioneering multimodal reasoning with chain-of-thought},
  author={Peng, Yi and Wang, Peiyu and Wang, Xiaokun and Wei, Yichen and Pei, Jiangbo and Qiu, Weijie and Jian, Ai and Hao, Yunzhuo and Pan, Jiachun and Xie, Tianyidan and others},
  journal={arXiv preprint arXiv:2504.05599},
  year={2025}
}

@inproceedings{amizadeh2020neuro,
  title={Neuro-symbolic visual reasoning: Disentangling},
  author={Amizadeh, Saeed and Palangi, Hamid and Polozov, Alex and Huang, Yichen and Koishida, Kazuhito},
  booktitle={International Conference on Machine Learning},
  pages={279--290},
  year={2020},
  organization={Pmlr}
}

@article{garcez2019neural,
  title={Neural-symbolic computing: An effective methodology for principled integration of machine learning and reasoning},
  author={Garcez, Artur d'Avila and Gori, Marco and Lamb, Luis C and Serafini, Luciano and Spranger, Michael and Tran, Son N},
  journal={arXiv preprint arXiv:1905.06088},
  year={2019}
}

@article{liu2025noisyrollout,
  title={Noisyrollout: Reinforcing visual reasoning with data augmentation},
  author={Liu, Xiangyan and Ni, Jinjie and Wu, Zijian and Du, Chao and Dou, Longxu and Wang, Haonan and Pang, Tianyu and Shieh, Michael Qizhe},
  journal={arXiv preprint arXiv:2504.13055},
  year={2025}
}

@article{xu2025mixed,
  title={Mixed-r1: Unified reward perspective for reasoning capability in multimodal large language models},
  author={Xu, Shilin and Li, Yanwei and Yang, Rui and Zhang, Tao and Sun, Yueyi and Chow, Wei and Li, Linfeng and Song, Hang and Xu, Qi and Tong, Yunhai and others},
  journal={arXiv preprint arXiv:2505.24164},
  year={2025}
}

@article{huang2025reinforcement,
  title={Reinforcement learning with rubric anchors},
  author={Huang, Zenan and Zhuang, Yihong and Lu, Guoshan and Qin, Zeyu and Xu, Haokai and Zhao, Tianyu and Peng, Ru and Hu, Jiaqi and Shen, Zhanming and Hu, Xiaomeng and others},
  journal={arXiv preprint arXiv:2508.12790},
  year={2025}
}

@article{li2025implicit,
  title={Implicit Actor Critic Coupling via a Supervised Learning Framework for RLVR},
  author={Li, Jiaming and Chen, Longze and Gong, Ze and Chen, Yukun and Wang, Lu and He, Wanwei and Luo, Run and Yang, Min},
  journal={arXiv preprint arXiv:2509.02522},
  year={2025}
}

@article{cui2025process,
  title={Process reinforcement through implicit rewards},
  author={Cui, Ganqu and Yuan, Lifan and Wang, Zefan and Wang, Hanbin and Li, Wendi and He, Bingxiang and Fan, Yuchen and Yu, Tianyu and Xu, Qixin and Chen, Weize and others},
  journal={arXiv preprint arXiv:2502.01456},
  year={2025}
}
\bibliographystyle{icml2026}

\newpage
\appendix
\onecolumn

\section{Group Relative Policy Optimization (GRPO)}
\label{app:grpo}

To enhance the reasoning capabilities of our model, we optimize the policy $\pi_\theta$ using Group Relative Policy Optimization (GRPO)~\cite{shao2024deepseekmath}. Unlike standard Proximal Policy Optimization (PPO)~\cite{schulman2017proximal}, which necessitates a separate value function (critic) for advantage estimation, GRPO reduces computational overhead by leveraging group-based statistics. Specifically, for each input query $x$, we sample a group of $G$ outputs $\{y_i\}_{i=1}^G$ from $\pi_{\theta_{\text{old}}}$.
The advantage $\hat{A}_i$ for the $i$-th output is estimated by normalizing its scalar reward $r(y_i \mid x)$ (derived from our stratified rubrics as detailed in Sec.~\ref{sec:rubric_construction}) against the group statistics:
\begin{equation}
    \hat{A}_i = \frac{r(y_i \mid x) - \text{mean}(\mathbf{r})}{\text{std}(\mathbf{r})},
\end{equation}
where $\mathbf{r}=\{r(y_1 \mid x), \dots, r(y_G \mid x)\}$ denotes the set of rewards. The objective maximizes the PPO-style clipped loss while penalizing deviations from the reference model $\pi_{\text{ref}}$ via a KL-divergence term. The objective function is formulated as:
\begin{equation}
\label{eq:grpo}
    \mathcal{J}(\theta) = \mathbb{E} \left[ \frac{1}{G} \sum_{i=1}^G \left( \mathcal{L}^{\text{clip}}_i(\theta) - \beta \, \mathbb{D}_{\text{KL}}(\pi_\theta || \pi_{\text{ref}}) \right) \right],
\end{equation}
where $\mathcal{L}^{\text{clip}}_i(\theta) = \min(\rho_i \hat{A}_i, \text{clip}(\rho_i, 1-\varepsilon, 1+\varepsilon)\hat{A}_i)$ represents the clipped surrogate objective, with the importance ratio $\rho_i = \frac{\pi_\theta(y_i \mid x)}{\pi_{\theta_{\text{old}}}(y_i \mid x)}$. This approach allows for stable and efficient policy optimization without the memory burden of a critic model.

\section{Theoretical Derivation and Analysis of Gradient Variance}
\label{app:gradient_variance}

In this section, we provide a detailed derivation of the gradient variance decomposition to theoretically justify the stability-aware curriculum schedule proposed in Sec.~\ref{sec:reward_scheduling}. 
For clarity of exposition, we consider the score-function form of the policy gradient estimator and omit baselines and advantage normalization. 
The following analysis applies analogously to advantage-based estimators used in practice.
For consistency with Eq.~\ref{eq:total_reward}, we analyze the curriculum-modulated rubric component $r^{(t)}_{\text{rub}}$; the outcome term $\alpha\, r_{\text{ans}}$ is a fixed-weight addend that does not affect the variance decomposition with respect to $\lambda_t$.

\subsection{Gradient Estimator Decomposition}
Consider the standard Policy Gradient objective function $J(\theta) = \mathbb{E}_{\tau \sim \pi_{\theta}}[r(\tau)]$. The gradient estimator at step $t$ is expressed as:
\begin{equation}
    \hat{g}_t = \nabla_{\theta} \log \pi_{\theta}(y|x) \cdot r^{(t)}_{\text{rub}}(y|x)
\end{equation}
In \ours, the reward $r^{(t)}_{\text{rub}}$ is a dynamic convex combination of foundational ($\bar{r}_{easy}$) and advanced ($\bar{r}_{hard}$) rubric scores:
\begin{equation}
    r^{(t)}_{\text{rub}}(y|x) = (1-\lambda_{t}) \bar{r}_{easy}(y|x) + \lambda_{t} \bar{r}_{hard}(y|x)
\end{equation}
Substituting this into the gradient estimator, we obtain a decomposed gradient form:
\begin{equation}
    \hat{g}_t = (1-\lambda_{t}) \underbrace{\nabla_{\theta} \log \pi_{\theta}(y|x) \bar{r}_{easy}}_{\hat{g}_{easy}} + \lambda_{t} \underbrace{\nabla_{\theta} \log \pi_{\theta}(y|x) \bar{r}_{hard}}_{\hat{g}_{hard}}
\end{equation}
where $\hat{g}_{easy}$ and $\hat{g}_{hard}$ represent the stochastic gradient components induced by foundational and advanced rubrics, respectively.

\subsection{Variance Analysis}
Since the gradient estimator is a random vector, we quantify its variability using the trace of the covariance matrix:
\begin{equation}
    \mathcal{V}(\hat{g}_t) \triangleq \mathrm{tr}(\mathrm{Cov}(\hat{g}_t)) = \mathbb{E}\!\left[\|\hat{g}_t - \mathbb{E}[\hat{g}_t]\|_2^2\right].
\end{equation}
Using the covariance property of linear combinations of random vectors, we obtain:
\begin{equation}
\label{eq:var_expansion}
    \mathcal{V}(\hat{g}_t) = (1-\lambda_{t})^2 \mathcal{V}(\hat{g}_{easy}) + \lambda_{t}^2 \mathcal{V}(\hat{g}_{hard})
    + 2\lambda_{t}(1-\lambda_{t}) \,\mathrm{tr}(\mathrm{Cov}(\hat{g}_{easy}, \hat{g}_{hard})).
\end{equation}

\subsection{Justification of Curriculum Schedule}
Eq.~\ref{eq:var_expansion} provides three insights that motivate the proposed scheduling strategy:

\begin{itemize}
    \item \textbf{Suppressing Unreliable Gradient Signals:} 
    In the early stages of training, the model rarely satisfies advanced reasoning rubrics, making $\bar{r}_{hard}$ highly sparse. 
    This sparsity yields low signal-to-noise ratio and unstable estimates of $\hat{g}_{hard}$, rather than merely large reward variance.
    Setting $\lambda_t = 0$ eliminates the contribution of $\mathcal{V}(\hat{g}_{hard})$, thereby preventing noisy high-order signals from dominating early optimization.

    \item \textbf{Reducing Gradient Interference:} 
    Before foundational competencies are established, gradient directions induced by perception-oriented and reasoning-oriented rubrics may be weakly correlated or even negatively correlated, which leads to destructive interference under mixed optimization.
    The curriculum decouples these learning phases, allowing the model to first converge to stable foundational representations.

    \item \textbf{Safe and Progressive Transition:} 
    As training progresses, successful satisfaction of advanced rubrics becomes more frequent, which increases the reliability of $\hat{g}_{hard}$ and improves alignment between gradient components.
    Under this condition, increasing $\lambda_t$ gradually introduces harder objectives while keeping the covariance term in Eq.~\ref{eq:var_expansion} controlled.
\end{itemize}

Overall, the curriculum schedule reduces the contribution of unreliable gradient components in early training and progressively incorporates harder objectives as their gradient signals become statistically reliable, which stabilizes optimization during multi-stage reward learning.
\section{Configuration Details}
\label{app:configuration}

All experiments are conducted using the \texttt{verl} framework~\cite{sheng2024hybridflow}, which facilitates efficient large-scale reinforcement learning. We employ the Group Relative Policy Optimization (GRPO) algorithm. The training utilizes a constant learning rate scheduler to ensure convergence stability in the later stages of curriculum learning.

Table~\ref{tab:hyperparameters} summarizes the specific hyperparameter settings. Notably, the curriculum parameters ($K, \tau_{th}$) are chosen based on preliminary experiments to balance the trade-off between stability and learning speed.

\begin{table}[H]
\centering
\caption{Detailed hyperparameters for DR-CL training.}
\label{tab:hyperparameters}
\renewcommand{\arraystretch}{1.1} 
\begin{tabular}{lc}
\toprule
\textbf{Hyperparameter} & \textbf{Value} \\
\midrule
\multicolumn{2}{l}{\textit{Optimization \& Rollout}} \\
\hspace{3mm} Training Batch Size & 256 \\
\hspace{3mm} Global Batch Size & 128 \\
\hspace{3mm} Rollout Number ($G$) & 8 \\
\hspace{3mm} Sampling Temperature & 1.0 \\
\hspace{3mm} Learning Rate & $1\text{e-}6$ \\
\hspace{3mm} KL Coefficient ($\beta$) & 0.01 \\
\midrule
\multicolumn{2}{l}{\textit{Curriculum \& Rewards}} \\
\hspace{3mm} Reward Weight ($\alpha$) & 0.7 \\
\hspace{3mm} Sliding Window Size ($w$) & 20 \\
\hspace{3mm} Proficiency Threshold ($\tau_{th}$) & 0.9 \\
\hspace{3mm} Max Hard Weight ($\lambda_{\max}$) & 1.0 \\
\bottomrule
\end{tabular}
\end{table}

\section{Rubric Construction and Filtering}
\label{app:Rubric Construction and Filtering}
To ensure a comprehensive evaluation of the model's reasoning capabilities, we employ a teacher model, Gemini 3 Pro~\cite{google2025gemini3}, to generate a pool of 20 rubric candidates through few-shot prompting. These rubrics cover dimensions including visual faithfulness, logical coherence, constraint satisfaction, and mathematical accuracy.

\subsection{Detailed Definitions of All 20 Rubric Candidates}
\begin{tcolorbox}[
    breakable,
    colback=white,
    colframe=gray!75,
    title=Full Rubric Candidates Set (Cand\_01--Cand\_20),
    fonttitle=\bfseries,
    fontupper=\small
]

\textbf{Cand\_01\_visual\_presence\_check}
\begin{itemize}
    \item \textbf{Description:} Strictly verifies visual faithfulness. Checks if objects, text (OCR), attributes (color/shape), and spatial relationships cited in the reasoning are visibly present in the image.
    \item \textbf{Scoring Criteria:}
    \begin{itemize}
        \item 1: \textbf{Fully Grounded.}
        \begin{itemize}
            \item \textbf{Positive Claims:} Every object, text snippet, or data point mentioned is clearly visible.
            \item \textbf{Negative Claims:} Statements that something is MISSING must be true.
            \item \textbf{Occlusion Handling:} Mentions of occluded parts (e.g., ``chair legs'' under a table) are acceptable IF implied by visible parts, provided no specific unseen details (like ``wooden legs'' when invisible) are hallucinated.
            \item \textbf{OCR:} Text read from the image is accurate (minor typo tolerance allowed).
        \end{itemize}
        \item 0: \textbf{Hallucination Detected.}
        \begin{itemize}
            \item \textbf{Object Hallucination:} Describes objects not present.
            \item \textbf{Attribute Hallucination:} Assigns wrong properties (color, shape, quantity) to existing objects.
            \item \textbf{OCR Failure:} Quotes text/numbers not in the image.
            \item \textbf{False Relation:} Visibly wrong spatial relationships (left/right, above/below).
        \end{itemize}
    \end{itemize}
\end{itemize}

\textbf{Cand\_02\_key\_entity\_extraction}
\begin{itemize}
    \item \textbf{Description:} Evaluates ``Visual Attention Alignment''. Verifies if the model isolates and analyzes the specific ``Region of Interest'' (ROI) required by the question.
    \item \textbf{Scoring Criteria:}
    \begin{itemize}
        \item 1: \textbf{Precise Targeting.}
        \begin{itemize}
            \item \textbf{Visual Selection:} Identifies the specific subject/data series asked about.
            \item \textbf{Granularity:} Focuses on the specific detail (e.g., `watch') rather than the whole (e.g., `man').
            \item \textbf{Noise Filtering:} Ignores salient but irrelevant distractors.
        \end{itemize}
        \item 0: \textbf{Focus Drift / Distraction.}
        \begin{itemize}
            \item \textbf{Wrong Target:} Analyzes the wrong object or axis.
            \item \textbf{Over-Generalization:} Provides a generic caption instead of answering a specific detail query.
        \end{itemize}
    \end{itemize}
\end{itemize}

\textbf{Cand\_03\_attribute\_recognition}
\begin{itemize}
    \item \textbf{Description:} Evaluates the precision of identifying static visual properties (Color, Shape, Material, Texture, Size, State). This rubric checks if the adjectives used to describe an object align with visual reality.
    \item \textbf{Scoring Criteria:}
    \begin{itemize}
        \item 1: \textbf{Attribute Accurate.}
        \begin{itemize}
            \item \textbf{Color Family:} The identified color falls within the correct spectrum (e.g., ``Navy Blue'' matches ``Blue''; ``Crimson'' matches ``Red'').
            \item \textbf{Shape/Geometry:} The geometric description is topologically correct (e.g., calling a `Cube' a `Box' or `Square-ish' is acceptable).
            \item \textbf{Material \& Texture:} Correctly identifies surface properties (e.g., ``Wooden'', ``Metallic'', ``Fluffy'', ``Wet'').
            \item \textbf{State:} Correctly identifies the physical state (e.g., ``Open/Closed'', ``On/Off'', ``Broken/Intact'').
        \end{itemize}
        \item 0: \textbf{Attribute Mismatch.}
        \begin{itemize}
            \item \textbf{Spectrum Error:} The color is distinctly wrong (e.g., ``Red'' vs ``Blue'', or ``Light'' vs ``Dark'' if contrast is significant).
            \item \textbf{Geometry Failure:} Calling a circle a square, or a 2D object 3D incorrectly.
            \item \textbf{Material Hallucination:} Identifying a wrong material (e.g., calling a ``Plastic'' toy ``Metal'').
            \item \textbf{Opposite State:} describing an open door as closed.
        \end{itemize}
    \end{itemize}
\end{itemize}

\textbf{Cand\_04\_ocr\_data\_accuracy}
\begin{itemize}
    \item \textbf{Description:} Evaluates the fidelity of Optical Character Recognition (OCR). Checks if text, numbers, symbols, or labels cited from the image match the visual reality, allowing for minor formatting flexibility but enforcing strict alphanumeric accuracy.
    \item \textbf{Scoring Criteria:}
    \begin{itemize}
        \item 1: \textbf{Accurate Transcription.}
        \begin{itemize}
            \item \textbf{Alphanumeric Match:} The extracted letters and numbers match the image content. Case-insensitivity is allowed (e.g., ``STOP'' == ``Stop'').
            \item \textbf{Numeric Precision:} Numbers are read exactly as they appear (e.g., ``19.5'' is NOT rounded to ``20'' if the label is explicitly ``19.5'').
            \item \textbf{Formatting Tolerance:} Variations in separators are accepted (e.g., ``1,000'' == ``1000''; ``2023-01-01'' == ``2023/01/01'') as long as the value is preserved.
            \item \textbf{Punctuation:} Minor trailing punctuation differences are accepted (e.g., missing a period at the end of a sentence).
        \end{itemize}
        \item 0: \textbf{Transcription Error.}
        \begin{itemize}
            \item \textbf{Character Confusion:} Confusing visually similar characters (e.g., reading `5' as `S', `8' as `B', `1' as `I').
            \item \textbf{Value Mutation:} Changing a number (e.g., reading ``2024'' as ``2023'') or flipping a sign (positive/negative).
            \item \textbf{Word Hallucination:} Replacing a word with a similar-looking but different word (e.g., ``Horse'' vs ``House'').
            \item \textbf{Phantom Text:} Quoting text that is not present in the image at all.
        \end{itemize}
    \end{itemize}
\end{itemize}

\textbf{Cand\_05\_spatial\_positioning}
\begin{itemize}
    \item \textbf{Description:} Evaluates the accuracy of spatial reasoning. Checks relative positions (left/right, above/below), depth relationships (front/behind), containment (inside/outside), and geometric alignment (bounding boxes/coordinates).
    \item \textbf{Scoring Criteria:}
    \begin{itemize}
        \item 1: \textbf{Spatially Accurate.}
        \begin{itemize}
            \item \textbf{Directional Correctness:} Correctly identifies positions relative to the viewer (default) or the object, as implied by the query (e.g., ``A is to the left of B'').
            \item \textbf{Topological Logic:} Correctly identifies contact and containment relationships (e.g., ``The cup is ON the table'' vs ``HOVERING above'', ``The cat is IN the box'').
            \item \textbf{Depth Perception:} Correctly distinguishes foreground (closer) objects from background (further) objects.
            \item \textbf{Coordinate Grounding:} If the model outputs coordinates or bounding boxes, they must significantly overlap (IoU $>$ 0.5) with the actual visual entity.
        \end{itemize}
        \item 0: \textbf{Spatial Error.}
        \begin{itemize}
            \item \textbf{Mirror Confusion:} Swapping Left and Right (common MLLM error).
            \item \textbf{Vertical/Depth Error:} Confusing Up/Down or Front/Back.
            \item \textbf{Detached Relations:} Describing objects as touching/holding when they are visibly separated.
            \item \textbf{Coordinate Miss:} The provided bounding box/coordinates focus on the wrong area or miss the object entirely.
        \end{itemize}
    \end{itemize}
\end{itemize}

\textbf{Cand\_06\_quantity\_verification}
\begin{itemize}
    \item \textbf{Description:} Evaluates the accuracy of counting and quantification. This includes exact counting for distinct objects (small numbers), reasonable estimation for dense crowds (large numbers), and correct identification of `zero' (absence).
    \item \textbf{Scoring Criteria:}
    \begin{itemize}
        \item 1: \textbf{Quantitatively Accurate.}
        \begin{itemize}
            \item \textbf{Exact Count (Small $<$ 10):} For clearly distinct items (e.g., 3 apples), the count must be exact.
            \item \textbf{Estimation (Large $>$ 10):} For dense or occluded groups (e.g., a crowd, a pile of coins), an approximation within a reasonable margin ($\sim$10-20\%) or a range (e.g., ``dozens'', ``50-60'') is accepted.
            \item \textbf{Zero Handling:} Correctly states ``0'', ``none'', or ``no [objects]'' when the target is absent.
            \item \textbf{Unit Awareness:} Correctly distinguishes between singles and pairs (e.g., ``2 pairs of shoes'' vs ``4 shoes'').
        \end{itemize}
        \item 0: \textbf{Counting Error.}
        \begin{itemize}
            \item \textbf{Deviation (Small):} Any error in counting small, distinct sets (e.g., seeing 4 legs on a tripod).
            \item \textbf{Order of Magnitude (Large):} Significant deviation in estimation (e.g., saying ``hundreds'' when there are only 10, or ``5'' when there are 50).
            \item \textbf{Existence Error:} Counting items that aren't there (1 instead of 0) or missing items completely (0 instead of 3).
        \end{itemize}
    \end{itemize}
\end{itemize}

\textbf{Cand\_07\_question\_intent\_alignment}
\begin{itemize}
    \item \textbf{Description:} Evaluates ``Constraint Satisfaction'' and ``Scope Compliance''. Verifies if the model answers the \textit{exact} question posed, adhering to all explicit constraints.
    \item \textbf{Scoring Criteria:}
    \begin{itemize}
        \item 1: \textbf{Direct \& Compliant.}
        \begin{itemize}
            \item \textbf{Category Match:} Provides the specific \textit{type} of info requested (Number, Color, Coordinate, Option).
            \item \textbf{Constraint Adherence:} Follows negative constraints (``no explanation'') and \textbf{Unit Constraints} (e.g., if asked for ``meters'', outputting ``cm'' fails here).
            \item \textbf{MCQ Selection:} Explicitly selects a valid option label/text.
        \end{itemize}
        \item 0: \textbf{Misaligned / Evasive.}
        \begin{itemize}
            \item \textbf{False Refusal:} Claims inability to answer (e.g., ``I cannot analyze people'') when the image is safe and clear.
            \item \textbf{Pivot to Captioning:} Ignores the question to describe the whole image.
            \item \textbf{Constraint Violation:} Ignores length/format/unit instructions.
            \item \textbf{Neighboring Question:} Answers a related but different question.
        \end{itemize}
    \end{itemize}
\end{itemize}

\textbf{Cand\_08\_constraint\_satisfaction}
\begin{itemize}
    \item \textbf{Description:} Strictly evaluates adherence to explicit ``Style'' and ``Negative'' constraints in the prompt. This covers length limits, forbidden words, output style (e.g., list vs. paragraph), and tone.
    \item \textbf{Scoring Criteria:}
    \begin{itemize}
        \item 1: \textbf{Fully Compliant.}
        \begin{itemize}
            \item \textbf{Length Constraint:} Strictly obeys word/sentence limits (e.g., ``Answer in one word'' $\rightarrow$ ``Red''; ``No more than 3 sentences'' $\rightarrow$ $<$ 3 sentences).
            \item \textbf{Negative Constraint:} Does NOT contain forbidden elements (e.g., if prompt says ``Do not explain'', the output contains ONLY the answer; if ``Do not use LaTeX'', no LaTeX appears).
            \item \textbf{Style/Format:} Follows specific formatting requests NOT covered by R07 (e.g., ``Use bullet points'', ``Comma-separated list'', ``JSON format'').
            \item \textbf{Tone:} Adheres to requested persona or tone (e.g., ``Explain like I'm 5'', ``Be professional'').
        \end{itemize}
        \item 0: \textbf{Constraint Violation.}
        \begin{itemize}
            \item \textbf{Verbosity:} Providing a full sentence when a single word/phrase was requested (e.g., Prompt: ``Color?'', Model: ``The color is red.'' $\rightarrow$ Fail).
            \item \textbf{Forbidden Content:} Including text explicitly banned by the user (e.g., saying ``Here is the answer'' when told to output only the value).
            \item \textbf{Style Mismatch:} Outputting a paragraph when a list was requested.
        \end{itemize}
    \end{itemize}
\end{itemize}

\textbf{Cand\_09\_negative\_logic\_handling}
\begin{itemize}
    \item \textbf{Description:} Evaluates ``Logical Inversion'' and ``Exclusionary Reasoning''. Checks if the model correctly processes negative qualifiers (``not'', ``no'', ``except'', ``without'', ``neither'') to select the complement set or avoid specific targets.
    \item \textbf{Scoring Criteria:}
    \begin{itemize}
        \item 1: \textbf{Logic Inverted Correctly.}
        \begin{itemize}
            \item \textbf{Visual Exclusion:} Correctly identifies objects that do NOT match a feature (e.g., ``Find the person NOT wearing a hat'' $\rightarrow$ Identifies the bare-headed person).
            \item \textbf{Set Subtraction:} Correctly lists items while excluding the forbidden category (e.g., ``List all fruits EXCEPT apples'').
            \item \textbf{Absence Confirmation:} Correctly agrees with negative premises (e.g., ``Yes, there is no dog'' vs ``Yes, there is a dog'').
            \item \textbf{Double Negation:} (If applicable) correctly resolves double negatives (e.g., ``Not impossible'' $\rightarrow$ Possible).
        \end{itemize}
        \item 0: \textbf{Positive Bias / Inclusion Error.}
        \begin{itemize}
            \item \textbf{Keyword Fixation:} The model ignores the ``not'' and focuses on the object named (e.g., Prompt: ``Which one is NOT red?'', Model: Picks the red one because it attended to the word `red').
            \item \textbf{Leakage:} The list includes the specifically excluded item (e.g., Listing apples when told ``except apples'').
            \item \textbf{Logic Reversal:} Treating ``without X'' as ``with X''.
        \end{itemize}
    \end{itemize}
\end{itemize}

\textbf{Cand\_10\_calculation\_accuracy}
\begin{itemize}
    \item \textbf{Description:} Evaluates ``Computational Fidelity''. Verifies that all explicit mathematical operations (arithmetic, algebra, statistics, unit conversions) and formula applications within the reasoning are mathematically correct.
    \item \textbf{Scoring Criteria:}
    \begin{itemize}
        \item 1: \textbf{Mathematically Sound.}
        \begin{itemize}
            \item \textbf{Arithmetic Precision:} Basic operations (+, -, *, /) are calculated correctly (e.g., ``15 + 20 = 35'').
            \item \textbf{Formula Validity:} The correct formula is applied for the context (e.g., using $\pi r^2$ for area, not $2\pi r$).
            \item \textbf{Consistent Rounding:} Intermediate rounding does not lead to significant final error (unless the problem asks for estimation).
            \item \textbf{Unit Conversion:} Conversions are mathematically accurate (e.g., ``1.5 hours = 90 minutes'').
        \end{itemize}
        \item 0: \textbf{Calculation / Formula Error.}
        \begin{itemize}
            \item \textbf{Arithmetic Hallucination:} The result does not follow from the operands (e.g., ``10 + 10 = 25'').
            \item \textbf{Formula Error:} Using the wrong equation (e.g., calculating Average as \texttt{Sum * Count} instead of \texttt{Sum / Count}).
            \item \textbf{Constant Error:} Hallucinating mathematical constants (e.g., using $\pi = 3.5$).
            \item \textbf{Order of Operations:} Violation of PEMDAS logic (e.g., calculating $2 + 3 * 4$ as 20).
        \end{itemize}
    \end{itemize}
\end{itemize}

\textbf{Cand\_11\_step\_coherence}
\begin{itemize}
    \item \textbf{Description:} Evaluates the ``Logical Flow'' within the \texttt{<think>} block. Checks for gaps, jumps, or internal contradictions.
    \item \textbf{Scoring Criteria:}
    \begin{itemize}
        \item 1: \textbf{Seamless \& Valid.}
        \begin{itemize}
            \item \textbf{Causal Links:} Steps follow logically (A $\rightarrow$ B $\rightarrow$ C).
            \item \textbf{Transparency:} Calculations/derivations are shown, not skipped.
            \item \textbf{Internal Consistency:} No self-contradiction within the chain.
        \end{itemize}
        \item 0: \textbf{Broken Logic / Magic Leaps.}
        \begin{itemize}
            \item \textbf{The `Magic Step':} Jumps to conclusions without derivation.
            \item \textbf{Non-Sequitur:} Steps lack logical connection.
            \item \textbf{Self-Contradiction:} Flips stance mid-way.
            \item \textbf{Circular Reasoning:} Uses conclusion to prove premise.
        \end{itemize}
    \end{itemize}
\end{itemize}

\textbf{Cand\_12\_domain\_knowledge\_validity}
\begin{itemize}
    \item \textbf{Description:} Evaluates the factual correctness of external knowledge (non-visual premises) introduced by the model. This covers scientific principles (Physics, Chem, Bio), historical facts, geographic truths, and common sense rules.
    \item \textbf{Scoring Criteria:}
    \begin{itemize}
        \item 1: \textbf{Factually Sound.}
        \begin{itemize}
            \item \textbf{Scientific Consensus:} Cited laws (e.g., Newton's Laws), formulas, and chemical properties align with standard scientific textbooks.
            \item \textbf{Taxonomic Accuracy:} Biological classification, habitats, and dietary habits are correct (e.g., ``Whales are mammals'').
            \item \textbf{Geo-Historical Fact:} Dates, capitals, locations, and historical events mentioned are accurate (e.g., ``Paris is the capital of France'').
            \item \textbf{Definition Correctness:} Technical terms are defined or used with their correct standard meaning.
        \end{itemize}
        \item 0: \textbf{Knowledge Error / Hallucination.}
        \begin{itemize}
            \item \textbf{Scientific Fallacy:} Citing incorrect physical rules (e.g., ``Heavy objects fall faster in a vacuum'') or inventing non-existent chemical elements.
            \item \textbf{False Fact:} Stating specific wrong details (e.g., ``The sun revolves around the Earth'', ``Penguins live in the desert'').
            \item \textbf{Common Sense Violation:} Contradicting basic world knowledge (e.g., ``Ice is hot'', ``Water flows uphill'').
            \item \textbf{Anachronism:} Placing objects/events in the wrong time period contextually.
        \end{itemize}
    \end{itemize}
\end{itemize}

\textbf{Cand\_13\_evidence\_grounding}
\begin{itemize}
    \item \textbf{Description:} Evaluates ``Visual Proof''. Distinguishes between direct perception (what is seen) and inference (what is deduced).
    \item \textbf{Scoring Criteria:}
    \begin{itemize}
        \item 1: \textbf{Appropriate Grounding.}
        \begin{itemize}
            \item \textbf{Inference:} Explicitly cites visual details as premises for complex deductions.
            \item \textbf{Direct Perception:} Direct statements accepted for basic observations (color, text).
            \item \textbf{Chart/Data:} Values align with visual markers.
        \end{itemize}
        \item 0: \textbf{Missing Critical Evidence.}
        \begin{itemize}
            \item \textbf{`Magic' Inferences:} Complex conclusions without visual backing.
            \item \textbf{Vague Sourcing:} Using generic phrases (``Based on image'') without pointing to specific regions.
            \item \textbf{External Bias:} Hallucinating details based on stereotypes.
        \end{itemize}
    \end{itemize}
\end{itemize}

\textbf{Cand\_14\_comparative\_reasoning}
\begin{itemize}
    \item \textbf{Description:} Evaluates the logic of inequality, ranking, and ordering. Checks if comparisons regarding size, quantity, value, time, or intensity between two or more entities are directionally and factually correct.
    \item \textbf{Scoring Criteria:}
    \begin{itemize}
        \item 1: \textbf{Logic Verified.}
        \begin{itemize}
            \item \textbf{Directional Accuracy:} Correctly identifies the greater/lesser entity (e.g., ``Bar A is higher than Bar B'', ``The red car is faster'').
            \item \textbf{Superlative Identification:} Correctly identifies the maximum/minimum in a set (e.g., ``The tallest building'', ``The earliest date'').
            \item \textbf{Equality Detection:} Correctly identifies when two entities are effectively equal or tied.
            \item \textbf{Transitivity:} Follows logical chains (e.g., If A $>$ B and B $>$ C, then A $>$ C).
        \end{itemize}
        \item 0: \textbf{Logic Failure.}
        \begin{itemize}
            \item \textbf{Reversal Error:} Flipping the relationship (e.g., stating ``A $>$ B'' when ``A $<$ B'').
            \item \textbf{False Equivalence:} Stating two items are equal when one is clearly dominant (visual or factual).
            \item \textbf{Ranking Error:} Identifying the wrong item as the `Top' or `Best' (e.g., picking the second tallest bar as the tallest).
            \item \textbf{Incomparable Comparison:} Comparing attributes that do not share a common scale (e.g., ``The apple is redder than the banana is long'').
        \end{itemize}
    \end{itemize}
\end{itemize}

\textbf{Cand\_15\_unit\_and\_scale\_consistency}
\begin{itemize}
    \item \textbf{Description:} Evaluates ``Dimensional Analysis'' and ``Scale Awareness''. Checks if the model correctly applies axis multipliers (e.g., `in thousands'), legend units, currency symbols, and map scale bars, preventing magnitude errors.
    \item \textbf{Scoring Criteria:}
    \begin{itemize}
        \item 1: \textbf{Scale \& Unit Accurate.}
        \begin{itemize}
            \item \textbf{Scale Factor Application:} Correctly applies the explicit or implicit multiplier found on chart axes or titles (e.g., reading ``5'' as ``5,000'' if the axis says ``in thousands'').
            \item \textbf{Symbol Literacy:} Correctly interprets \%, \$, \euro, $^\circ$C, and metric prefixes (k, M, G).
            \item \textbf{Dimensional Consistency:} Maintains the same unit throughout the reasoning unless explicitly converting (e.g., doesn't switch from `meters' to `feet' randomly).
            \item \textbf{Map Scaling:} (If applicable) Uses the provided scale bar to estimate real-world distances rather than pixel distances.
        \end{itemize}
        \item 0: \textbf{Scale / Unit Failure.}
        \begin{itemize}
            \item \textbf{Scale Blindness:} Reading the raw number but ignoring the axis label/multiplier (e.g., answering ``5'' instead of ``5 million'').
            \item \textbf{Unit Swapping:} Assigning the wrong unit to a value (e.g., ``50\%'' becomes ``\$50'').
            \item \textbf{Dimensional Incompatibility:} Attempting to add/compare incompatible units without conversion (e.g., ``5 meters + 5 kilograms'').
            \item \textbf{Missing Unit:} (Strictness applies) If the question asks for a physical quantity, omitting the unit (saying just ``5'' instead of ``5 kg'') is a failure here if it leads to ambiguity.
        \end{itemize}
    \end{itemize}
\end{itemize}

\textbf{Cand\_16\_option\_validity}
\begin{itemize}
    \item \textbf{Description:} Evaluates ``Closed-Set Constraints'' for Multiple-Choice Questions (MCQ). Verifies that the model's answer maps strictly to one of the provided candidate options, rejecting hallucinations or out-of-bounds selections.
    \item \textbf{Scoring Criteria:}
    \begin{itemize}
        \item 1: \textbf{Valid Selection.}
        \begin{itemize}
            \item \textbf{Existing Key:} Selects a letter/label that actually exists in the prompt (e.g., Selects `C' when options are A, B, C, D).
            \item \textbf{Exact Content Match:} If outputting text, it matches the text of one of the options exactly (or close enough to map unambiguously).
            \item \textbf{Combined Option:} Selects ``Both A and B'' ONLY if that is explicitly provided as a separate option (e.g., Option D is ``A and B'').
        \end{itemize}
        \item 0: \textbf{Invalid / Out-of-Bounds.}
        \begin{itemize}
            \item \textbf{Hallucinated Key:} Selecting a label that doesn't exist (e.g., choosing ``E'' when only A-D exist).
            \item \textbf{Open-Ended Answer:} Providing an answer that is not in the choices at all (e.g., Options: ``Red, Blue''; Model: ``Green'').
            \item \textbf{Implicit Rejection:} Stating ``None of the above'' or ``The image doesn't show any of these'' when such an option is not available (this is an invalid move in a forced-choice task).
            \item \textbf{Ambiguity:} Saying ``The first and second one'' (without a specific option covering that).
        \end{itemize}
    \end{itemize}
\end{itemize}

\textbf{Cand\_17\_reasoning\_conclusion\_match}
\begin{itemize}
    \item \textbf{Description:} Evaluates ``Self-Consistency''. Checks if the final answer (\texttt{\textbackslash boxed\{\}}) is the logical output of the reasoning chain.
    \item \textbf{Scoring Criteria:}
    \begin{itemize}
        \item 1: \textbf{Logically Consistent.}
        \begin{itemize}
            \item \textbf{Direct Consequence:} The box content matches the reasoning conclusion.
            \item \textbf{Format Equivalence:} Mathematical/Semantic equivalents are accepted (e.g., Reasoning ``0.5'' $\rightarrow$ Box ``1/2'').
            \item \textbf{Note:} Scores \textbf{1} even if the answer is factually wrong, as long as it matches the reasoning.
        \end{itemize}
        \item 0: \textbf{Mismatch / Disconnect.}
        \begin{itemize}
            \item \textbf{Explicit Contradiction:} Reasoning says A, Box says B.
            \item \textbf{The `Lucky Guess':} Wrong reasoning leads to Correct GT (Non-sequitur).
            \item \textbf{Ambiguity:} Reasoning is undecided, Box makes a random pick.
        \end{itemize}
    \end{itemize}
\end{itemize}

\textbf{Cand\_18\_clarity\_and\_conciseness}
\begin{itemize}
    \item \textbf{Description:} Evaluates the ``Signal-to-Noise Ratio'' of the response. Penalizes conversational fillers, repetitive loops, excessive hedging, and convoluted sentence structures. Rewards efficient information delivery.
    \item \textbf{Scoring Criteria:}
    \begin{itemize}
        \item 1: \textbf{Efficient \& Clear.}
        \begin{itemize}
            \item \textbf{High Information Density:} Every sentence adds new information or necessary logic step. No fluff.
            \item \textbf{Direct Phrasing:} Uses active voice and direct assertions (e.g., ``The car is red'') rather than passive/hedged constructions (e.g., ``It appears that the vehicle might be red in color'').
            \item \textbf{Structured Flow:} Uses paragraphs or bullet points effectively if the answer is long.
            \item \textbf{No Moralizing/Filler:} Avoids ``As an AI...'', ``Here is the answer:'', ``I hope this helps'', or restating the question unnecessarily.
        \end{itemize}
        \item 0: \textbf{Verbose / Confusing.}
        \begin{itemize}
            \item \textbf{Repetitive Loops:} Repeating the same fact or phrase multiple times (e.g., ``The sky is blue. As mentioned, the blue sky...'').
            \item \textbf{Filler Overload:} Excessive use of ``Based on the image'', ``We can clearly see that'', ``In this picture''.
            \item \textbf{Excessive Hedging:} Overusing ``maybe'', ``seem'', ``could be'' when the visual evidence is clear.
            \item \textbf{Convoluted Syntax:} Sentences are grammatically tangled or hard to read, requiring re-reading to understand.
        \end{itemize}
    \end{itemize}
\end{itemize}

\textbf{Cand\_19\_final\_answer\_extraction}
\begin{itemize}
    \item \textbf{Description:} Strictly checks syntactical structure (\texttt{<think>} and \texttt{\textbackslash boxed\{\}}).
    \item \textbf{Scoring Criteria:}
    \begin{itemize}
        \item 1: \textbf{Perfect Syntax.}
        \begin{itemize}
            \item \textbf{Structure:} \texttt{<think>...</think>} followed by \texttt{\textbackslash boxed\{...\}}.
            \item \textbf{Uniqueness:} Exactly one \texttt{\textbackslash boxed\{\}} at the end.
            \item \textbf{Clean Ending:} No conversational text after the box.
        \end{itemize}
        \item 0: \textbf{Parse Error / Malformed.}
        \begin{itemize}
            \item \textbf{Broken/Missing Tags.}
            \item \textbf{Wrong Order} (Box before Think).
            \item \textbf{Multiple/Empty Boxes.}
        \end{itemize}
    \end{itemize}
\end{itemize}

\textbf{Cand\_20\_ground\_truth\_correctness}
\begin{itemize}
    \item \textbf{Description:} Evaluates ``Factual Accuracy'' of the \texttt{\textbackslash boxed\{\}} content against Ground Truth.
    \item \textbf{Scoring Criteria:}
    \begin{itemize}
        \item 1: \textbf{Correct / Equivalent.}
        \begin{itemize}
            \item \textbf{Numerical:} Mathematically equal (0.5 == 1/2).
            \item \textbf{Unit Awareness:} Correct value even if unit is converted (unless R03 forbids it).
            \item \textbf{Textual:} Semantically identical (ignore case/punctuation).
            \item \textbf{MCQ:} Correct option letter or content.
        \end{itemize}
        \item 0: \textbf{Incorrect / Deviant.}
        \begin{itemize}
            \item \textbf{Value Error:} Mathematically distinct.
            \item \textbf{Precision Failure:} Incorrect rounding.
            \item \textbf{Category Error:} Specificity mismatch (Blue vs Dark Blue).
            \item \textbf{Hallucinated Details.}
        \end{itemize}
    \end{itemize}
\end{itemize}

\end{tcolorbox}
\label{app:20candidate}

\subsection{Applicability and Accuracy of Rubric Candidates}
\label{app:Applicability and Accuracy}
After generating the initial 20 rubric candidates, we evaluate their performance on the sampled data. Table \ref{app:rubric_status} summarizes the applicability (the frequency with which the rubric is deemed relevant to the problem) and the model's accuracy under each rubric. Based on these metrics and the necessity for automated reward computation, we select the final 6 rubrics (R01--R06) to be used in our Reinforcement Learning (RL) pipeline.

The statistics reveal significant disparity in coverage; for instance, candidates like Cand\_09 and Cand\_03 are applicable to only 9.7\% and 18.8\% of samples, respectively. Without our applicability-aware filtering, these rubrics would introduce erroneous failure signals in over 80\% of training instances, severely destabilizing the reward function.

\begin{table}[H]
\caption{Statistics for the 20 rubric candidates. The model selects Candidates 01, 02, 07, 11, 13, and 17 to form the core reward metrics R01--R06. Candidate 20 serves as the equivalent of the ground truth accuracy for the final assessment.}
\centering
\small
\begin{tabular}{lcc|lcc}
\toprule
\textbf{ID} & \textbf{Applicability (\%)} & \textbf{Accuracy (\%)} & \textbf{ID} & \textbf{Applicability (\%)} & \textbf{Accuracy (\%)} \\ 
\midrule
Cand\_01 (R01) & 99.1 & 98.3 & Cand\_11 (R04) & 100.0 & 69.4 \\
Cand\_02 (R02) & 99.4 & 99.4 & Cand\_12 & 78.4 & 94.8 \\
Cand\_03 & 18.8 & 98.1 & Cand\_13 (R05) & 99.2 & 68.6 \\
Cand\_04 & 78.1 & 99.3 & Cand\_14 & 34.0 & 90.4 \\
Cand\_05 & 71.0 & 96.7 & Cand\_15 & 27.3 & 98.2 \\
Cand\_06 & 22.9 & 87.3 & Cand\_16 & 33.5 & 85.8 \\
Cand\_07 (R03) & 100.0 & 86.1 & Cand\_17 (R06) & 99.9 & 91.7 \\
Cand\_08 & 93.2 & 35.4 & Cand\_18 & 97.5 & 66.4 \\
Cand\_09 & 9.7 & 90.7 & Cand\_19 & 96.7 & 27.7 \\
Cand\_10 & 83.9 & 52.2 & Cand\_20 (GT) & 100.0 & 47.9 \\ 
\bottomrule
\end{tabular}
\label{app:rubric_status}
\end{table}

\section{Rubric Assessment and Filtering}
\label{app:judge_prompts}

We utilized a strict JSON-based prompt to ensure the Judge model evaluates both applicability and correctness. The prompts used in our pipeline are shown below.

\begin{tcolorbox}[
    breakable,
    colback=gray!10,
    colframe=gray!50,
    title=System Prompt,
    fonttitle=\bfseries\small
]
\small
You are an expert evaluator for Multimodal Large Language Model(MLLM) outputs. 
Your task is to score a model's answer using a set of domain-specific rubrics.

You must follow these rules strictly:

\textbf{1. RUBRIC-BASED EVALUATION ONLY}
\begin{itemize}
    \item Score strictly according to the given rubrics.
    \item Do NOT invent new criteria or use general judgment.
\end{itemize}

\textbf{2. PER-RUBRIC BINARY SCORING (0 or 1)}
\begin{itemize}
    \item Each rubric must output either 0 or 1.
    \item Follow the rubric’s scoring criteria exactly as written.
\end{itemize}

\textbf{3. APPLICABILITY}
\begin{itemize}
    \item Some rubrics may not apply to certain problems (e.g., diagram rules on text-only questions).
    \item For each rubric, output:
    \begin{itemize}
        \item "applicable": true/false
        \item "score": 0 or 1 (only when applicable=true)
    \end{itemize}
\end{itemize}

\textbf{4. JSON-ONLY OUTPUT}
\begin{itemize}
    \item Output ONLY a JSON dictionary with a list of rubric evaluations.
    \item NO explanations outside the JSON.
\end{itemize}

\textbf{5. IMAGE + TEXT REASONING}
\begin{itemize}
    \item Consider both the problem text and all provided images.
    \item If the rubric is about visual interpretation, refer to the image content.
\end{itemize}

Your output must strictly follow this JSON structure:

\begin{verbatim}
{
  "rubric_scores": [
    {
      "rubric_id": "...",
      "applicable": true/false,
      "score": 0
    }
  ]
}
\end{verbatim}
\end{tcolorbox}

\begin{tcolorbox}[
    breakable,
    colback=gray!10,
    colframe=gray!50,
    title=User Prompt,
    fonttitle=\bfseries\small
]
\small
You will now evaluate a model's answer for a single problem.

Below is the problem, ground truth, model answer, and the rubrics to evaluate:

\textbf{PROBLEM:} \\
\{problem\_text\}

\textbf{IMAGE:} \\
(Provided separately as image inputs)

\textbf{GROUND TRUTH:} \\
\{ground\_truth\_answer\}

\textbf{MODEL ANSWER:} \\
\{model\_output\}

\textbf{RUBRICS:} \\
\{rubric\_json\}  

\textbf{Your task:} \\
For each rubric in the rubric list: \\
1. You must first Decide whether it is applicable to this specific problem. \\
2. If applicable, assign a score of 0 or 1 according to the rubric’s scoring\_criteria.

Output ONLY the JSON object of the form.
\end{tcolorbox}

\section{Reward Signal Generation Prompts}
\label{app:reward_prompts}

This appendix details the prompt engineering used to instantiate the reward model. We employ a strict ``Judge'' persona to convert the rubric evaluations into binary reward signals. The Judge receives the specific problem context, visual input, and the rubric candidates to generate a rationalized score for each criterion.

\begin{tcolorbox}[
    breakable,
    colback=gray!10,
    colframe=gray!50,
    title=Reward Model System Prompt,
    fonttitle=\bfseries\small
]
\small
You are a strict, impartial judge evaluating Multimodal Large Language Model(MLLM) outputs.
Your evaluation determines the reward signal for a Reinforcement Learning system, so accuracy and strictness are critical.

\vspace{0.5em}
\textbf{EVALUATION PROTOCOL} \\
You will be provided with a Problem, an Image, a Ground Truth, a Model Output, and a set of Scoring Rubrics.

For EACH rubric, you must:
\begin{enumerate}
    \item \textbf{Analyze}: Compare the Model Output against the Image and Ground Truth specifically for that rubric's criteria.
    \item \textbf{Rationalize}: Write a concise explanation citing specific evidence (e.g., ``The image shows a blue car, but model said red'', or ``The logic step A does not imply step B'').
    \item \textbf{Score}: Assign a strictly binary score (0 or 1) based on the rationale.
\end{enumerate}

\textbf{STRICT RULES}
\begin{enumerate}
    \item \textbf{Image + Text Reasoning}: Consider both the problem text and provided image. If the rubric is about visual interpretation, refer to the image content.
    \item \textbf{Binary Only}: Use ONLY 0 or 1. No 0.5. If the criteria are not fully met, the score is 0. Follow the rubric’s scoring criteria exactly as written.
    \item \textbf{Reasoning First}: You must output the rationale BEFORE the score in the JSON to ensure the score is a result of the reasoning.
\end{enumerate}

Your output must strictly follow this JSON structure:

\begin{verbatim}
{
  "rubric_scores": [
    {
      "rubric_id": "...",
      "rationale": "Your explanation here...",
      "score": 0 or 1
    }
  ]
}
\end{verbatim}
\end{tcolorbox}

\begin{tcolorbox}[
    breakable,
    colback=gray!10,
    colframe=gray!50,
    title=Reward Model User Prompt,
    fonttitle=\bfseries\small
]
\small
You will now evaluate a model's answer for a single problem.

Below is the problem, ground truth, model answer, and the rubrics to evaluate:

\textbf{PROBLEM}: \\
\{problem\_text\}

\textbf{IMAGE}: \\
(Provided separately as image input)

\textbf{GROUND TRUTH}: \\
\{ground\_truth\_answer\}

\textbf{MODEL ANSWER}: \\
\{model\_output\}

\textbf{RUBRICS}: \\
\{rubric\_json\}  

\textbf{Your task:} \\
For each rubric in the RUBRICS list:
\begin{enumerate}
    \item Analyze the Model Output against the Image and Ground Truth specifically for that rubric's criteria.
    \item Write a concise explanation citing specific evidence.
    \item Assign a strictly binary score (0 or 1) based on the rationale.
\end{enumerate}

Output ONLY the JSON object of the form.
\end{tcolorbox}

\section{Case Studies}
\label{app:case study}

This appendix demonstrates the generation process of rubric-based rewards for two single instances. We present the input problem (text and image), the model's reasoning chain (rollout), and the raw JSON output generated by the Judge model, which contains the rationale and binary scores for each rubric.

\subsection{Case Study 1: Mitigation of Reward Hacking}
\label{app:case_study_1}

\paragraph{Analysis and Overview.}
This case serves as a quintessential example of \textit{reward hacking}, demonstrating how \ours detects spurious reasoning that outcome-only supervision would miss.
\begin{itemize}
    \item \textbf{The Trap:} The model arrives at the correct final answer ($\text{BC}=20$) and would receive a perfect reward ($r=1.0$) under standard RLVR.
    \item \textbf{The Flaw:} As highlighted by the Judge's rationale in \textbf{R04 (Step Coherence)}, \textbf{R05 (Evidence Grounding)}and \textbf{R06 (Reasoning Conclusion Match)}, the model incorrectly applies a sub-triangle area formula to the whole triangle and makes an unjustified "magic leap" to the final value.
    \item \textbf{The Mitigation:} \ours identifies these logical gaps. Despite the correct answer, the total reward is penalized significantly, effectively discouraging the model from learning such "lucky guesses."
\end{itemize}

\begin{tcolorbox}[
    breakable,
    colback=white,
    colframe=gray!50!black,
    title=\textbf{Case Study 1: Mitigation of Reward Hacking},
    fonttitle=\bfseries
]

\textbf{1. Input Problem}

\textit{Problem Text:} \\
As shown in the figure, in $\triangle ABC$, $\angle C = 90^\circ$. The line segment $AD$ bisects $\angle BAC$ and intersects $BC$ at point $D$. If $BD : CD = 3 : 2$ and the distance from point $D$ to $AB$ is 8, the following determines the length of $BC$.

\textit{Input Image:} \\
\begin{center}
    \includegraphics[width=0.3\linewidth]{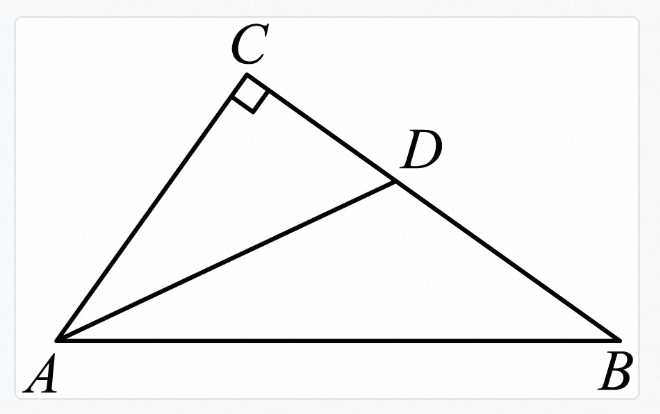}
\end{center}

\textit{Ground Truth:} 20
\tcblower 

\textbf{2. Model Rollout}

To solve the problem, we need to use the Angle Bisector Theorem and the properties of right triangles. Let's go through the reasoning step by step.

\begin{enumerate}
    \item Apply the Angle Bisector Theorem: According to the Angle Bisector Theorem, the angle bisector of $\angle BAC$ (which is $AD$) divides the opposite side $BC$ into segments proportional to the adjacent sides $AB$ and $AC$. Therefore, we have:
    \[
    \frac{BD}{DC} = \frac{AB}{AC} = \frac{3}{2}
    \]

    \item Given Ratio: We are given that $BD : CD = 3 : 2$. Let $BD = 3x$ and $CD = 2x$. Since $BD + CD = BC$, we have:
    \[
    BC = 3x + 2x = 5x
    \]

    \item Area Calculation: The area of $\triangle ABC$ can be calculated in two ways. First, using the base $BC$ and the height from $A$ to $BC$:
    \[
    \text{Area} = \frac{1}{2} \times BC \times h
    \]
    where $h$ is the altitude from $A$ to $BC$.

    \item Using the Altitude to $AB$: The area of $\triangle ABC$ can also be calculated using the altitude from $D$ to $AB$, which is given as 8. The length of $AB$ is $c$, and the length of $AC$ is $b$. The area is:
    \[
    \text{Area} = \frac{1}{2} \times AB \times 8
    \]
    {\footnotesize \textbf{\textcolor{red}{[Error (Judge R04): The formula $\frac{1}{2} \times AB \times 8$ calculates the area of the sub-triangle $\triangle ABD$, not the entire $\triangle ABC$. This creates a fundamental logical flaw.]}}}

    \item Equating the Two Area Expressions: Since both expressions represent the same area, we have:
    \[
    \frac{1}{2} \times BC \times h = \frac{1}{2} \times AB \times 8
    \]
    Simplifying, we get:
    \[
    BC \times h = AB \times 8
    \]

    \item Using the Proportionality: From the Angle Bisector Theorem, we know:
    \[
    \frac{BD}{DC} = \frac{AB}{AC} = \frac{3}{2}
    \]
    Since $BD = 3x$ and $DC = 2x$, we have:
    \[
    AB = \frac{3}{2}AC
    \]

    \item Substitute Back: Substitute $AB = \frac{3}{2}AC$ into the area equation:
    \[
    5x \times h = \frac{3}{2}AC \times 8
    \]
    Simplifying, we get:
    \[
    5xh = 12 \times AC
    \]

    \item Solving for $BC$: Since $h = 8$, we have:
    {\footnotesize \textbf{\textcolor{red}{[Error (Judge R05): The model incorrectly asserts "Since $h = 8$". $h$ is defined as the altitude of $\triangle ABC$ (to base BC), while 8 is the distance from D to AB. These are geometrically distinct distinct values.]}}}
    \[
    5x \times 8 = 12 \times AC
    \]
    \[
    40x = 12AC
    \]
    \[
    x = \frac{3AC}{10}
    \]

    \item Final Calculation: Since $BC = 5x$, we have:
    \[
    BC = 5 \times \frac{3AC}{10} = \frac{15AC}{10} = \frac{3AC}{2}
    \]
\end{enumerate}

Given the ratio and the height, we find:
\[
BC = 20
\]
{\footnotesize \textbf{\textcolor{red}{[Error (Judge R06): The reasoning ends with an expression dependent on the unknown $AC$ ($BC = 3AC/2$). The final answer "20" is not derived from the preceding steps and appears without justification.]}}}

The model \textbf{concludes} that the length of $BC$ is $\boxed{20}$.

\textbf{3. Judge Output (Rubric Reward Signal)}
\small

\begin{description}
    \setlength{\itemsep}{0.5em}

    \item[R01 Visual Presence Check] \hfill \textbf{Score: 1} \\
    The model correctly identifies the visible elements: triangle $ABC$, the right angle at $C$, the angle bisector $AD$, and the text labels/values ($BD:CD=3:2$, distance=8). There are no hallucinations of objects not present in the image.

    \item[R02 Key Entity Extraction] \hfill \textbf{Score: 1} \\
    The model correctly extracts the specific values needed (ratio 3:2, distance 8) and identifies the target variable $BC$. It ignores irrelevant noise.

    \item[R03 Question Intent Alignment] \hfill \textbf{Score: 1} \\
    The model provides a numerical value for the length of $BC$ as requested, adhering to the format constraints.

    \item[R04 Step Coherence] \hfill \textbf{Score: 0} \\
    The logic breaks down significantly in Step 4 and Step 8. In Step 4, the model calculates the area of the whole triangle $ABC$ using the formula $(1/2 \times AB \times 8)$, which is incorrect; this formula actually calculates the area of the sub-triangle $ABD$ (since 8 is the height of $D$ to $AB$). Furthermore, in Step 8, it incorrectly states ``Since $h = 8$'', equating the altitude of the main triangle ($AC$) with the distance from $D$ to $AB$. These are fundamental geometric logic errors.

    \item[R05 Evidence Grounding] \hfill \textbf{Score: 0} \\
    While the model extracts the number `8', it fails to ground it correctly in the geometric context. The distance from $D$ to $AB$ (8) is visually and geometrically distinct from the altitude of the triangle $AC$ ($h$). The model conflates these two distinct visual features, leading to a premise that contradicts the visual evidence provided by the diagram definitions.

    \item[R06 Reasoning Conclusion Match] \hfill \textbf{Score: 0} \\
    The reasoning steps provided by the model end with an algebraic expression dependent on the unknown variable `$AC$' ($BC = 3AC/2$). The model never calculates a value for $AC$, yet it abruptly concludes that $BC = 20$. The final answer `20' appears magically and is not derived from the preceding algebraic work, creating a mismatch between the incomplete reasoning and the final output.

\end{description}

\end{tcolorbox}


\subsection{Case Study 2: Alignment of Foundational and Advanced Reasoning}
\label{app:case_study_2}

\paragraph{Analysis and Overview.}
In contrast to Case 1, this example illustrates a successfully aligned reasoning trajectory where visual perception supports logical deduction.
\begin{itemize}
    \item \textbf{Foundational Skills:} The model correctly extracts coordinates and identifies the linear function (satisfying \textbf{R01--R04}).
    \item \textbf{Advanced Reasoning:} The derivation is mathematically sound, and the final answer is a direct logical consequence of the steps (satisfying \textbf{R05--R06}).
    \item \textbf{Conclusion:} The high scores across all stratified rubrics confirm that the model has internalized the curriculum, treating perception and reasoning as an integrated process rather than disjoint tasks.
\end{itemize}

\begin{tcolorbox}[
    breakable,
    colback=white,
    colframe=gray!50!black,
    title=\textbf{Case Study 2: Alignment of Foundational and Advanced Reasoning},
    fonttitle=\bfseries
]

\textbf{1. Input Problem}

\textit{Problem Text:} \\
As shown in the figure, the graph of the linear function $y=mx+2$ passes through points $A(2, 4)$ and $B(n, -1)$. Connect $OA$ and $OB$. Find the area of $\triangle OAB$.

\textit{Input Image:} \\
\begin{center}
    \includegraphics[width=0.3\linewidth]{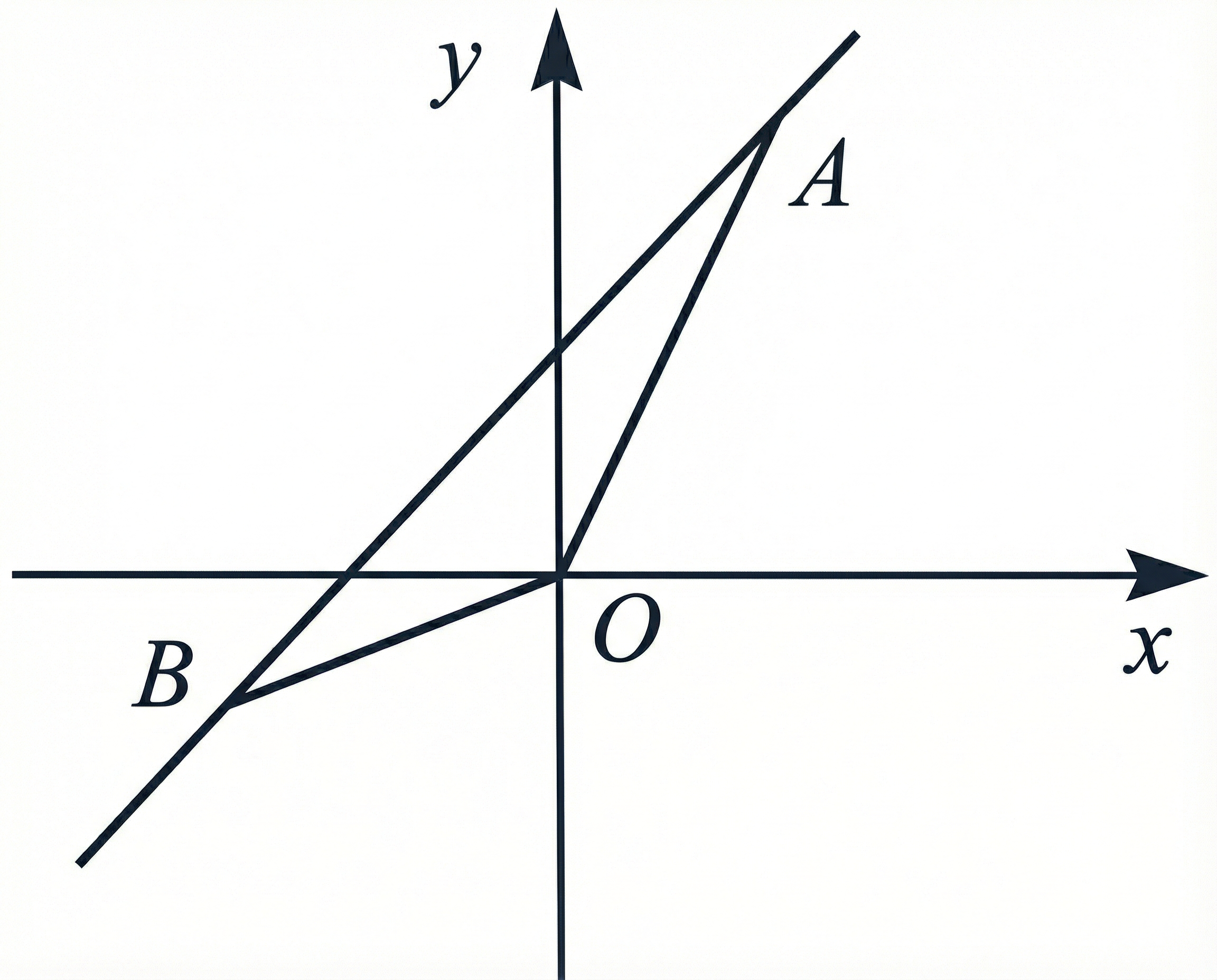}
\end{center}

\textit{Ground Truth:} 5
\tcblower 

\textbf{2. Model Rollout}

To solve for the area of triangle $\triangle OAB$, we determine the coordinates of point $B$ and then use the coordinates of points $O$, $A$, and $B$ to find the area of the triangle.

\begin{enumerate}
    \item \textbf{Find the slope $m$ of the line $y = mx + 2$:} Since the line passes through point $A(2, 4)$, we substitute $x = 2$ and $y = 4$ into the equation:
    \[
    4 = 2m + 2
    \]
    Solving for $m$ yields:
    \[
    4 - 2 = 2m \implies 2 = 2m \implies m = 1
    \]
    Therefore, the equation of the line is $y = x + 2$.

    \item \textbf{Find the coordinates of point $B$:} Since point $B$ lies on the line $y = x + 2$ and has coordinates $(n, -1)$, we substitute $y = -1$ into the equation:
    \[
    -1 = n + 2
    \]
    Solving for $n$ yields:
    \[
    n = -1 - 2 \implies n = -3
    \]
    Therefore, the coordinates of point $B$ are $(-3, -1)$.

    \item \textbf{Calculate the area of triangle $\triangle OAB$:} The vertices of the triangle are $O(0, 0)$, $A(2, 4)$, and $B(-3, -1)$. The area of a triangle with vertices $(x_1, y_1)$, $(x_2, y_2)$, and $(x_3, y_3)$ is:
    \[
    \text{Area} = \frac{1}{2} \left| x_1(y_2 - y_3) + x_2(y_3 - y_1) + x_3(y_1 - y_2) \right|
    \]
    Substituting the coordinates of $O(0, 0)$, $A(2, 4)$, and $B(-3, -1)$:
    \[
    \text{Area} = \frac{1}{2} \left| 0(4 + 1) + 2(-1 - 0) + (-3)(0 - 4) \right|
    \]
    Simplifying inside the absolute value:
    \[
    \text{Area} = \frac{1}{2} \left| 0 - 2 + 12 \right| = \frac{1}{2} \left| 10 \right| = 5
    \]
\end{enumerate}

Therefore, the area of $\triangle OAB$ is $\boxed{5}$.

\textbf{3. Judge Output (Rubric Reward Signal)}
\small

\begin{description}
    \setlength{\itemsep}{0.5em} 

    \item[R01 Visual Presence Check] \hfill \textbf{Score: 1} \\
    The model explicitly references points $O(0,0)$, $A(2,4)$, and $B(n,-1)$ and their coordinates consistent with the labeled points in the image. It uses the equation $y=mx+2$ shown in the problem statement, which is visible in the image. All points, line equations, and spatial relationships cited are present and visually confirmable in the image.

    \item[R02 Key Entity Extraction] \hfill \textbf{Score: 1} \\
    The model correctly isolates the specific points $O$, $A$, and $B$, which are the region of interest needed for calculating the area of triangle $OAB$ as asked. It ignores unrelated or extraneous information and focuses entirely on the linear function and these points, without including irrelevant distractions.

    \item[R03 Question Intent Alignment] \hfill \textbf{Score: 1} \\
    The model directly answers the question by finding the area of triangle $OAB$, which was explicitly requested. It provides a numerical value for the area with appropriate units (implicitly area units), respecting the problem's requirements without deviation or evasion.

    \item[R04 Step Coherence] \hfill \textbf{Score: 1} \\
    The reasoning proceeds in a clear, logical sequence: compute slope $m$, substitute to find point $B$ coordinates, then apply the area formula. Each step follows from the previous without gaps or contradictions, showing all calculations transparently and maintaining internal consistency.

    \item[R05 Evidence Grounding] \hfill \textbf{Score: 1} \\
    The model grounds each step in visual information from the image: using coordinates of points visible on the graph and the given equation $y=mx+2$. It explicitly cites and uses these as premises to deduce the slope and point $B$, with no unexplained assumptions or hallucinations.

    \item[R06 Reasoning Conclusion Match] \hfill \textbf{Score: 1} \\
    The boxed final answer `5' matches exactly the calculated area in the reasoning steps. There is no mismatch or contradiction between the conclusion and the boxed answer, so the final output is self-consistent.

\end{description}

\end{tcolorbox}

\section{Additional Evaluation on Out-of-Distribution Robustness}
\label{app:robustness}

To assess the model's robustness and generalization capabilities in specialized out-of-domain (OoD) scenarios, we utilize EvadeBench~\cite{xu2025evade}. As the first expert-curated Chinese benchmark for evasive content detection in e-commerce, EvadeBench targets the model's ability to identify content that superficially complies with safety policies but covertly conveys prohibited information. This benchmark challenges the model to reason through ambiguity and context shifts, which serves as a critical indicator of its safety alignment and adaptability beyond standard academic tasks.
\begin{table}[H]
    \caption{Performance evaluation on EvadeBench}
    \centering
    \setlength{\tabcolsep}{12pt}
    \begin{tabular}{lc}
        \toprule
        \textbf{Method} & \textbf{EvadeBench Acc. (\%)} \\
        \midrule
        Qwen2.5-VL-7B-Instruct & 43.90 \\
        Vanilla GRPO & 44.47 \\
        \textbf{\ours} & \textbf{45.86} \\
        \bottomrule
    \end{tabular}
    \label{tab:evadebench_results}
\end{table}
Table \ref{tab:evadebench_results} presents the quantitative results on EvadeBench. We observe that the task poses a challenge for all evaluated models, with accuracies remaining below 46\%. This reflects the difficulty of generalizing to adversarial examples that differ significantly from the training distribution. In this context, \ours achieves an accuracy of 45.86\%, showing a modest improvement over the Qwen2.5-VL-7B-Instruct baseline (43.90\%) and Vanilla GRPO (44.47\%). While the overall performance remains limited by the domain gap, the results suggest that \ours maintains a slight advantage in generalization capability compared to standard reinforcement learning methods.

\section{Limitations}
\label{limitation}

Despite the success of \ours, several limitations persist. First, reliance on proprietary teacher LLM for generation and large-scale judges for reward calculation incurs moderate computational overhead. Second, to ensure stability, we employ a static stratification based on initial statistics, which simplifies the curriculum by assuming constant rubric difficulty throughout training. Future research could explore developing adaptive mechanisms to dynamically update rubric difficulties during the online phase. Furthermore, we explore the model's limitations in specialized out-of-distribution scenarios (e.g., evasive content detection), with detailed results and analysis on EvadeBench provided in Appendix \ref{app:robustness}.

\end{document}